\documentclass[conference]{IEEEtran}
\IEEEoverridecommandlockouts
\usepackage{cite}
\usepackage{amsmath,amssymb,amsfonts}
\usepackage{algorithmic}
\usepackage{graphicx}
\usepackage{textcomp}
\usepackage{xcolor}
\usepackage{caption}
\usepackage{subcaption}
\def\BibTeX{{\rm B\kern-.05em{\sc i\kern-.025em b}\kern-.08em
    T\kern-.1667em\lower.7ex\hbox{E}\kern-.125emX}}
\begin{document}

\title{Exploring the Impact of Hand Pose and Shadow on Hand-washing Action Recognition
\thanks{This research was partially supported by the Indiana State Department of Agriculture Specialty Crop Grant Program \#A337-21-SCBG-20-103.}
}

\author{\IEEEauthorblockN{Shengtai Ju, Amy R. Reibman}
\IEEEauthorblockA{\textit{Elmore Family School of Electrical and Computer Engineering} \\
\textit{Purdue University}\\
West Lafayette, USA \\
ju10@purdue.edu, reibman@purdue.edu}
}

\maketitle
\begin{abstract}
In the real world, camera-based application systems can face many challenges, including environmental factors and distribution shift. In this paper, we investigate how pose and shadow impact a classifier's performance, using the specific application of handwashing action recognition. To accomplish this, we generate synthetic data with desired variations to introduce controlled distribution shift. Using our synthetic dataset, we define a classifier's breakdown points to be where the system's performance starts to degrade sharply, and we show these are heavily impacted by pose and shadow conditions. In particular, heavier and larger shadows create earlier breakdown points. Also, it is intriguing to observe model accuracy drop to almost zero with bigger changes in pose. Moreover, we propose a simple mitigation strategy for pose-induced breakdown points by utilizing additional training data from non-canonical poses. Results show that the optimal choices of additional training poses are those with moderate deviations from the canonical poses with 50-60 degrees of rotation.  


\end{abstract}    
\section{Introduction}
\label{sec:intro}
When designing application systems that leverage neural networks and deep learning, it is common to train and evaluate performance by splitting a large dataset into training and testing sets. This is based on the assumption that both training and testing sets come from the same underlying distribution. However, neural networks can be susceptible to small changes or distribution shifts in test data \cite{recht2019imagenet}. Change in poses of objects is one source of distribution shift, which can impact system performance for tasks like object classification \cite{strikewithapose} and person re-ID \cite{sun2019reid}. However, there is a lack of quantitative studies of poses in action recognition.
Therefore, in this paper, we explore the role that pose plays in the task of action recognition, specifically focusing on handwashing action recognition for food safety \cite{ju2023robust, zhong2019hand, zhong2020multi}.

In a food-handling setting, handwashing is closely associated with the spread of food-borne illnesses. According to the World Health Organization (WHO), food-borne illnesses sickens 600 million and even causes 420,000 deaths every year \cite{who_food}. Good hand hygiene practices are essential for promoting public welfare. In this paper, we follow the WHO handwashing guidelines \cite{who_guideline}, as shown in Figure~\ref{fig:who_step}. 

Compared to handwashing in medical settings, food-related handwashing is often conducted outdoors, where environmental conditions are more pronounced, primarily due to the presence of shadow. Because of this, designing a food-related handwashing recognition system becomes more challenging. Designing a system that is robust to lighting variation is therefore essential compared to indoor medical applications. In our previous work \cite{ju2023robust}, we have shown that model robustness is heavily impacted by changes in environmental conditions, including shadow conditions across different datasets. Therefore, in addition to the impact of hand poses, we also investigate the impact of various shadow conditions.

\begin{figure}[htb!]
\centering
\begin{subfigure}[t]{0.45\linewidth}
    \includegraphics[width=\linewidth]{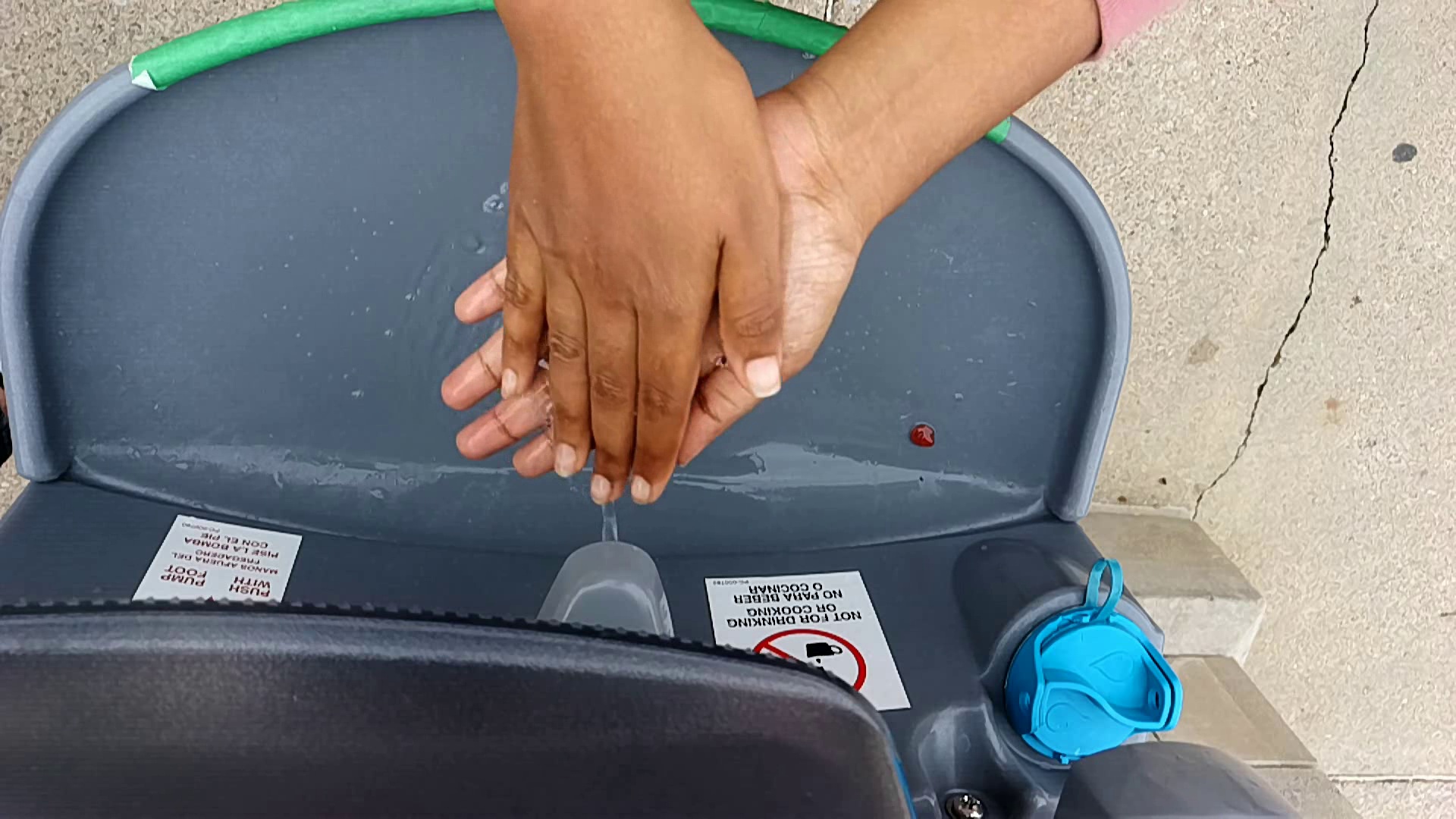}
    \caption{Rub palm to palm canonical angle.}
    \label{fig:rp_ca}
\end{subfigure}
\hspace{0.1cm}
\begin{subfigure}[t]{0.45\linewidth}
    \includegraphics[width=\linewidth]{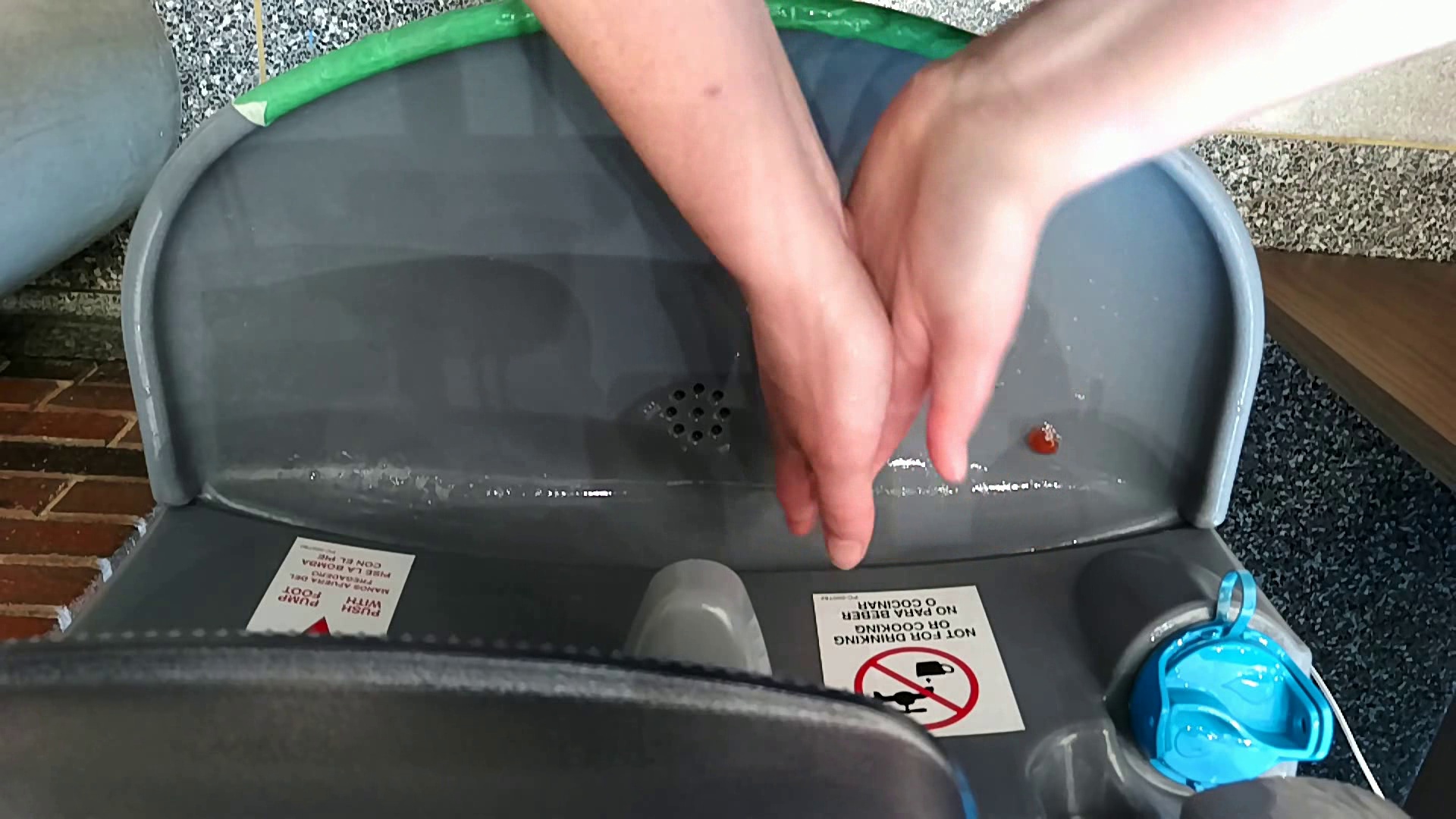}
    \caption{Rub palm to palm non-canonical angle.}
    \label{fig:rp_nonca}
\end{subfigure}
\vspace{.2cm}
\begin{subfigure}[b]{0.45\linewidth}
    \includegraphics[width=\linewidth]{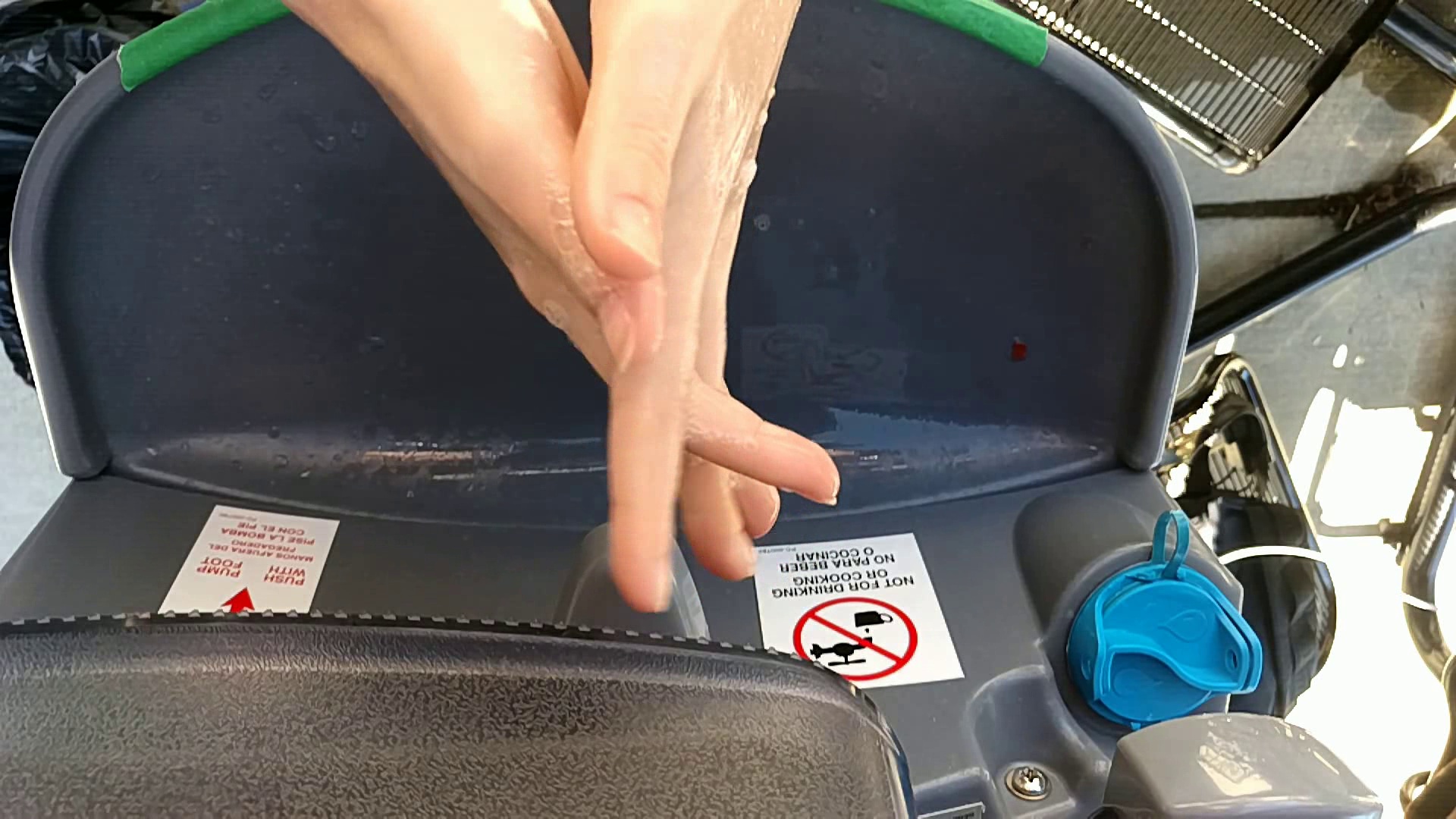}
    \caption{Rub palm with fingers interlaced canonical angle.}
    \label{fig:rpfi_ca}
\end{subfigure}
\hspace{0.1cm}
\begin{subfigure}[b]{0.45\linewidth}
    \includegraphics[width=\linewidth]{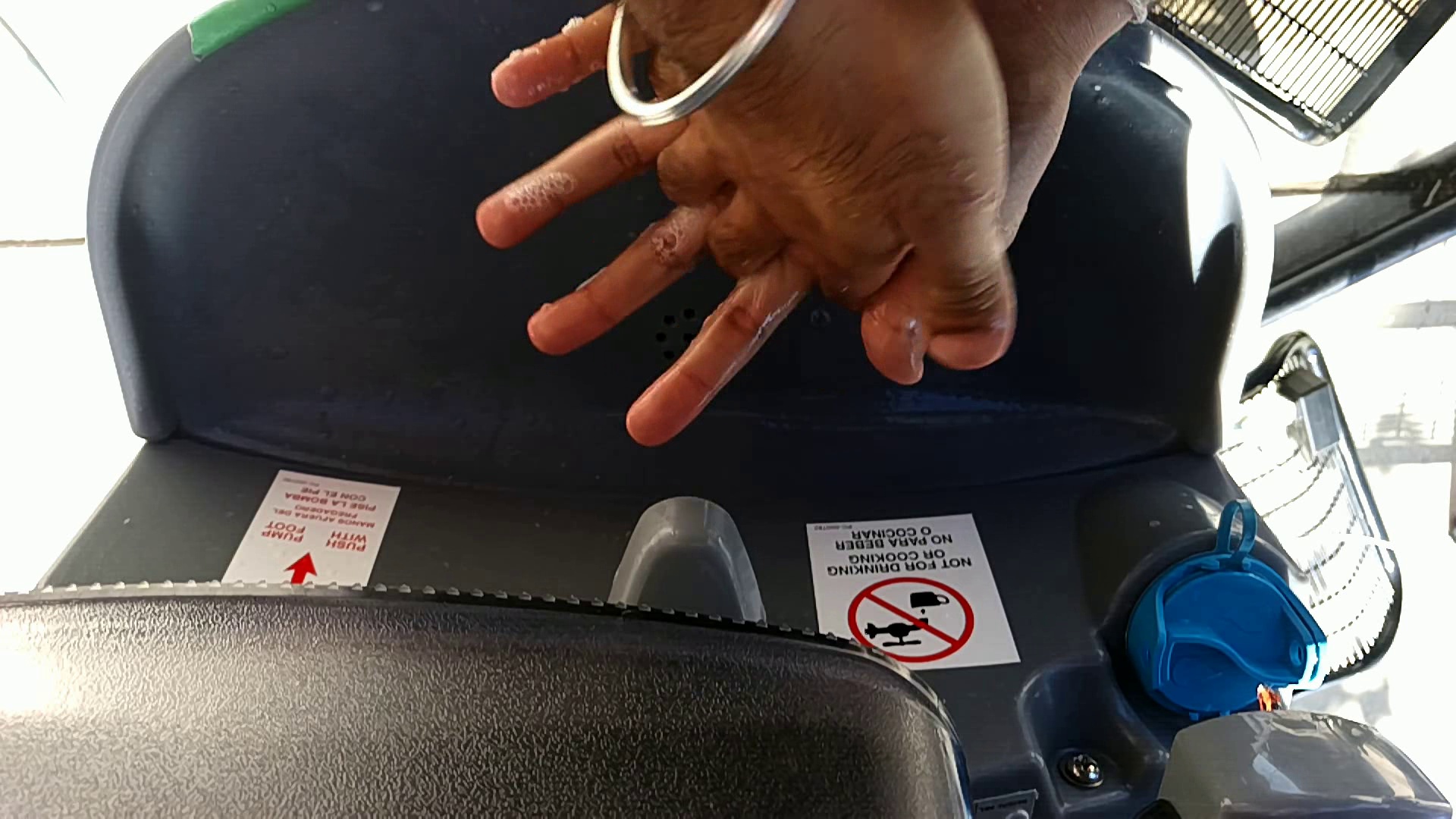}
    \caption{Rub palm with fingers interlaced non-canonical angle.}
    \label{fig:rpfi_nonca}
\end{subfigure}
        
\caption{Examples of different hand poses.}
\label{fig:ca_nonca_ex}
\end{figure}

Figure~\ref{fig:ca_nonca_ex} illustrates the diversity of hand poses seen from the dataset in \cite{ju2023robust}. Each row contains two images of the same action but in different poses. 
Most people perform the handwashing actions in canonical poses when following the given instructions. However, people may find the canonical poses uncomfortable, or they may perform the actions in non-canonical poses unconsciously, causing a distribution shift from the canonical poses. In these cases, a recognition system that has only been trained on the canonical poses could fail significantly when trying to recognize other poses. To investigate and better understand these challenges, we introduce and perform in-depth analysis of the breakdown points of a handwashing recognition system in terms of hand poses.



To investigate the impact of hand poses and shadow, it is necessary to collect image or video data that encompasses the desired variations. Two approaches can be taken for data collection: real-world data recording or synthetic data generation. The advantages of using real-world recording is that data is realistic and contains many natural variations that synthetic data cannot reproduce. However, to explore the impact of hand pose and shadow specifically, it would be necessary to estimate hand angles and the amount of shadow from the captured videos. The estimates of these data attributes can be inaccurate, and collecting enough data with sufficient desired variations would be labor intensive. With synthetic data, however, we will have accurate labels of all data attributes including poses and shadow. 

In the literature, synthetic datasets have been explored and proven to be effective ways to alleviate the labor intensive process of capturing data and generating ground truth labels \cite{varol2017learning, liu2019learning}. 
By purposefully constructing a dataset that contains the desired variations, we can obtain reliable and convincing results more efficiently. Therefore, in this paper, we have chosen to generate synthetic data to produce accurate and controllable parameters.

The main contributions of our work can be summarized as, 
\begin{enumerate}
    \item Quantitatively study to what extent hand poses impact a recognition system’s performance;
    \item Evaluate a handwashing action recognition system using breakdown points; 
    \item Provide an in-depth analysis of how shadow impacts a system's recognition performance using various shadow intensities, sizes, translations, and rotations; and
    \item Provide insights for future research on dataset collection and construction. 
\end{enumerate}

\section{Background and Related Work}
\label{sec:related_work}
\subsection{Handwashing Steps}
\label{subsec: who_steps}

\begin{figure}[htb!]
\centering
\includegraphics[width=\linewidth]{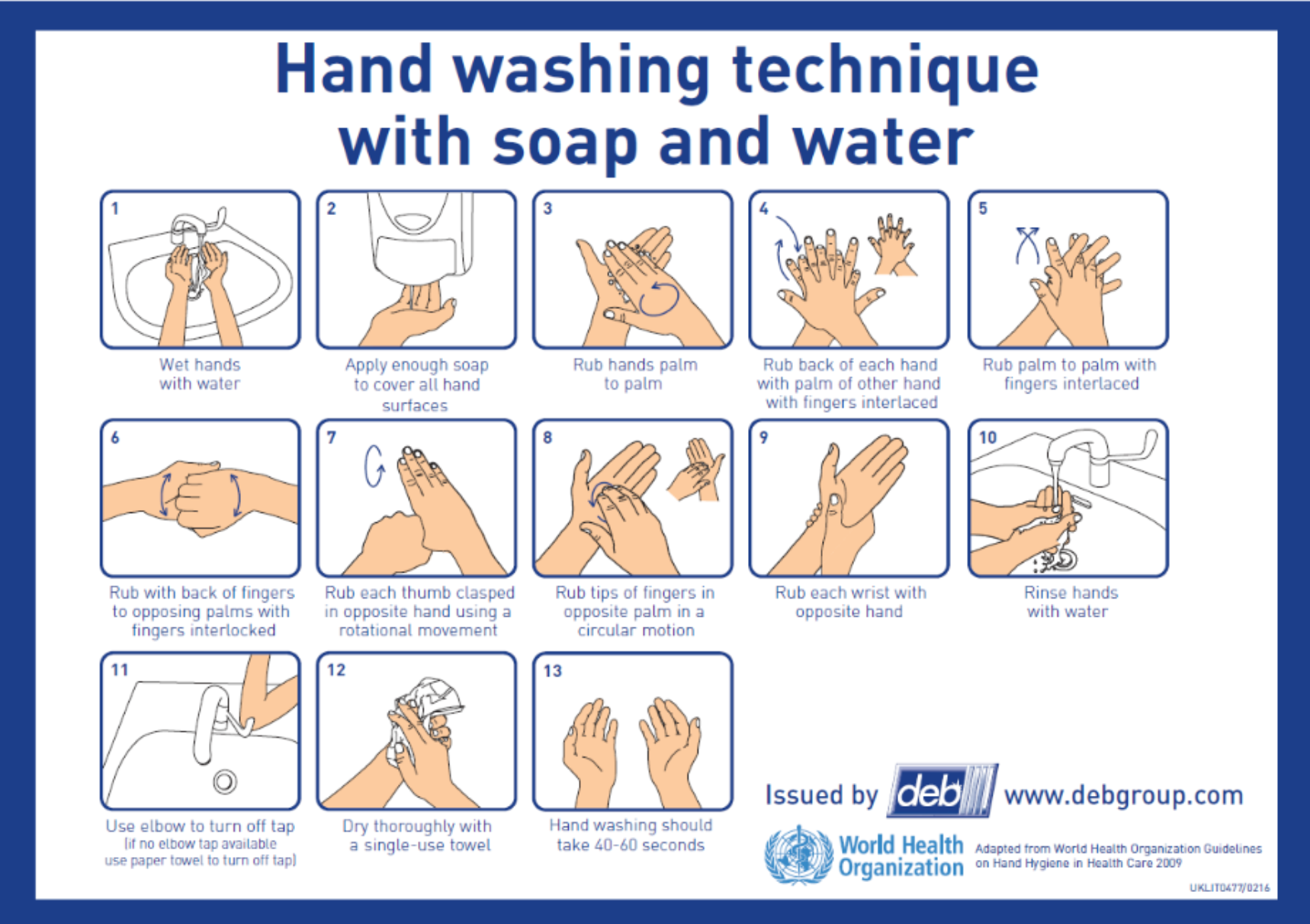}
\caption{The WHO Handwashing Steps.}
\label{fig:who_step}
\end{figure}

As shown in Figure~\ref{fig:who_step}, the WHO handwashing guidelines consist of 12 steps, of which 7 are rubbing actions and 5 are non-rubbing actions. The rubbing actions are the most prominent steps as they ensure all surfaces of hands are properly cleaned. Among the 7 rubbing actions, the hand poses of rubbing back of fingers with fingers interlocked and rubbing each wrist make them difficult and much more uncomfortable to be performed at different angles. Therefore, we exclude steps 6 and 9 from our study and further focus on the rubbing actions that can easily be performed in different angles. 

\subsection{Existing Handwashing Datasets}
Choices of datasets are crucial for evaluating a handwashing recognition system. Using real-world data, it has been shown that model performance drops as the evaluation environments become more complex \cite{elsts2022cnn}. Also, as shown in our previous work \cite{ju2023robust}, handwashing action recognition accuracy drops by more than 20\% when evaluated under different shadow conditions. 
There are public real-world datasets available for handwashing recognition that follow the WHO steps. For example, public datasets such as \cite{kaggle}, \cite{latvia_data}, and \cite{chinese_journal}
are useful for evaluating a system's robustness to domain changes and distribution shift in the real world as they contain enough variations across datasets. However, they have all been captured indoors with limited lighting/shadow variations, and they lack accurate instances of different hand angles we want to study. 

\subsection{Systems for Handwashing Recognition}

In hospitals and healthcare settings, handwashing recognition has been widely explored. 
A handwashing tutorial system for hospitals was developed through a series of works \cite{lacey_system1, lacey_system2, lacey_system3}. 
More recently, deep-learning-based systems for handwashing were proposed \cite{wang2022handwashing, chinese_journal, li2022hand_keyactionscore}. These works have shown promising results of designing and applying a vision-based system for recognizing the more challenging rubbing actions by WHO. However, they have only considered applications in hospitals or healthcare environments, whereas we are focusing on outdoor scenarios in food-safety. 


In the context of handwashing for safe food-handling, several works have explored methods to recognize the various actions. For egocentric handwashing videos, a two stream network that combines the RGB and optical flow streams was proposed by \cite{zhong2019hand}. This system was further extended into a two-stage system \cite{zhong2020multi} and a multi-room and multi-camera environment \cite{zhong2021designing}. Though these works have considered applying handwashing recognition for safer food-handling, they have not considered attributes such as pose and shadow which could significantly impact the system's performance. 



\subsection{Image Classification and Action Recognition}
\label{subsec:classifier}

Image classification and action recognition have been popular areas of research with many well-known models. For the task of image classification, CNN model architectures including VGG \cite{vgg}, DenseNet \cite{densenet}, ResNet \cite{resnet}, and the more recent VisionTransformer \cite{vit} have all demonstrated great performance on the benchmarking ImageNet dataset \cite{imagenet}. In addition, action recognition is the task of classifying actions present in a video to a set of predefined classes. Models including \cite{c3d, i3d, TRN, TSM, vivit, videoswin} have been proposed. These methods have shown impressive results on benchmarking action recognition datasets \cite{kinetics, sthsth}.


Given computational constraints and the various shadow settings we consider, our animated synthetic handwashing actions are limited to 20 frames for each action. The temporal length of our data, compared to other real-world video datasets such as \cite{kinetics} and \cite{sthsth}, is relatively short, thus making it less suitable for video-based classifiers. Therefore, we choose to use an image classifier for all experiments. Considering the complexity of our data and the application scenario, which requires fast and real-time feedback, we have chosen to use the lightweight MobileNetV3 \cite{mbnetv3} model as our classifier rather than more complex models.

\section{Dataset}
\label{sec:dataset}
In this section, we discuss the process of synthesizing our hand data with controllable parameters. We also discuss in detail about the software, models, and reasoning for our data.


\subsection{Actions}

Figure~\ref{fig:who_step} shows the detailed WHO handwashing steps. As mentioned earlier in Section~\ref{subsec: who_steps}, we only focus on the 7 rubbing actions (steps 3-9) which are more relevant for action recognition. Moreover, we exclude steps 6 and 9 because human hands cannot be easily rotated as in other actions. The 5 rubbing actions and our shortened names are: 
\begin{enumerate}
    \item Rub hands palm to palm (rub palm);
    \item Rub back of each hand with palm of other hand (rub back);
    \item Rub palm to palm with fingers interlaced (rub fingers interlaced);
    \item Rub each thumb with opposite hand (rub thumb); and
    \item Rub tips of fingers in opposite palm (rub tips).
\end{enumerate}

\subsection{Synthetic Data Creation (Without Shadow)}
\begin{figure}[htb!]
\centering
\begin{subfigure}[t]{0.31\linewidth}
    \includegraphics[width=\textwidth]{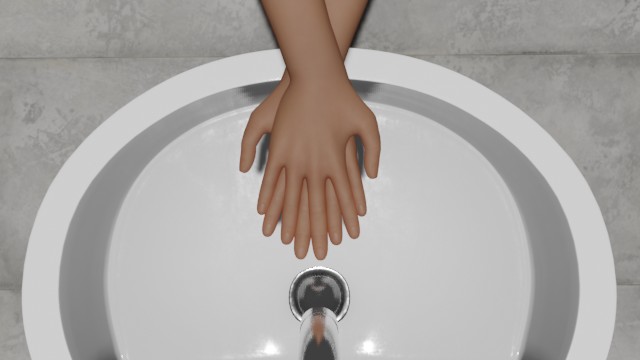}
    \caption{Rub back.}
    \label{fig:rb}
\end{subfigure}%
\hspace{.1cm}
\begin{subfigure}[t]{0.31\linewidth}
    \includegraphics[width=\textwidth]{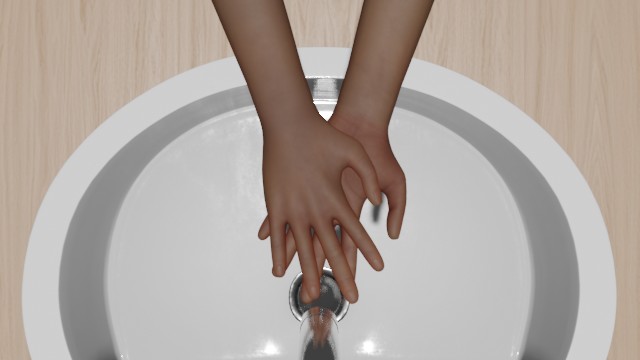}
    \caption{Rub palm.}
    \label{fig:rp}
\end{subfigure}%
\hspace{0.1cm}
\begin{subfigure}[t]{0.31\linewidth}
    \includegraphics[width=\textwidth]{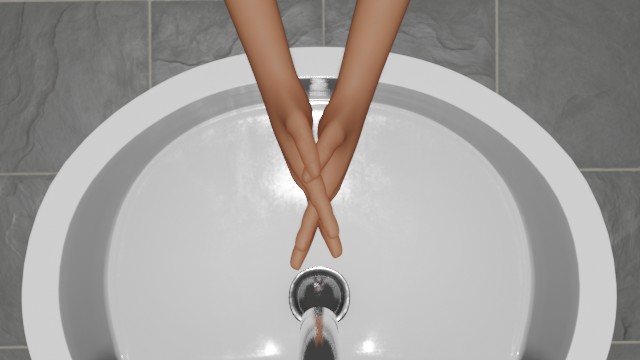}
    \caption{Rub fingers interlaced.}
    \label{fig:rpfi}
\end{subfigure}
\vspace{.1cm}
\begin{subfigure}[t]{0.31\linewidth}
    \includegraphics[width=\textwidth]{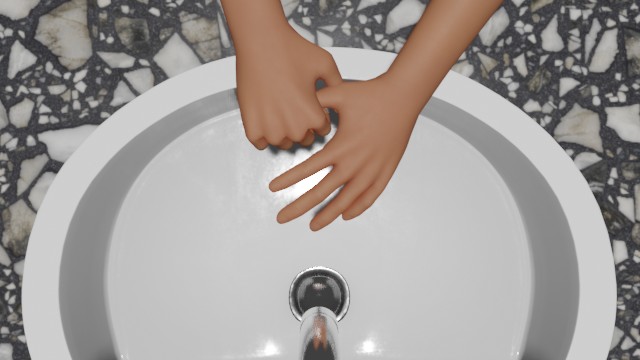}
    \caption{Rub thumb.}
    \label{fig:rth}
\end{subfigure}
\hspace{.1cm}
\begin{subfigure}[t]{0.31\linewidth}
    \includegraphics[width=\textwidth]{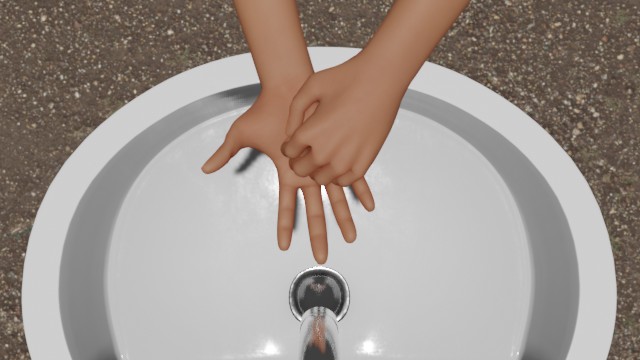}
    \caption{Rub tips 1.}
    \label{fig:rtips1}
\end{subfigure}
\hspace{.1cm}
\begin{subfigure}[t]{0.31\linewidth}
    \includegraphics[width=\textwidth]{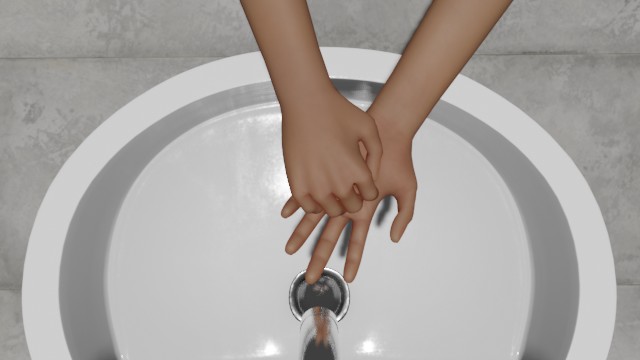}
    \caption{Rub tips 2.}
    \label{fig:rtips2}
\end{subfigure}%
\caption{Examples of hand poses.}
\label{fig:action_ex}
\end{figure}

For generating synthetic hand data, we use the free 3D software Blender for creating and rendering hand images. We purchased a high-quality hand model with realistic skin textures and fully adjustable joints. Then we started the data creation process by manually constructing hand images that match the 5 rubbing actions by carefully adjusting hand poses and finger joints. Next, we animated each action using their corresponding motion pattern. For example, the actions of rub palm and rub tips involve circular motions, and the actions of rub back and rub fingers interlaced involve back and forth motion. Figure~\ref{fig:action_ex} shows an example for each of the 5 rubbing actions. All actions except for rub fingers interlaced have a dual pose, which is constructed by interchanging the order of hands. For example, the action of rub tips can be done with left fingertips or right fingertips, as shown in Figures~\ref{fig:rtips1} and~\ref{fig:rtips2}. Similarly, the dual poses for other actions can be constructed in the same manner. 

\begin{figure}[htb!]
\centering
\begin{subfigure}[t]{0.22\linewidth}
    \includegraphics[width=\textwidth]{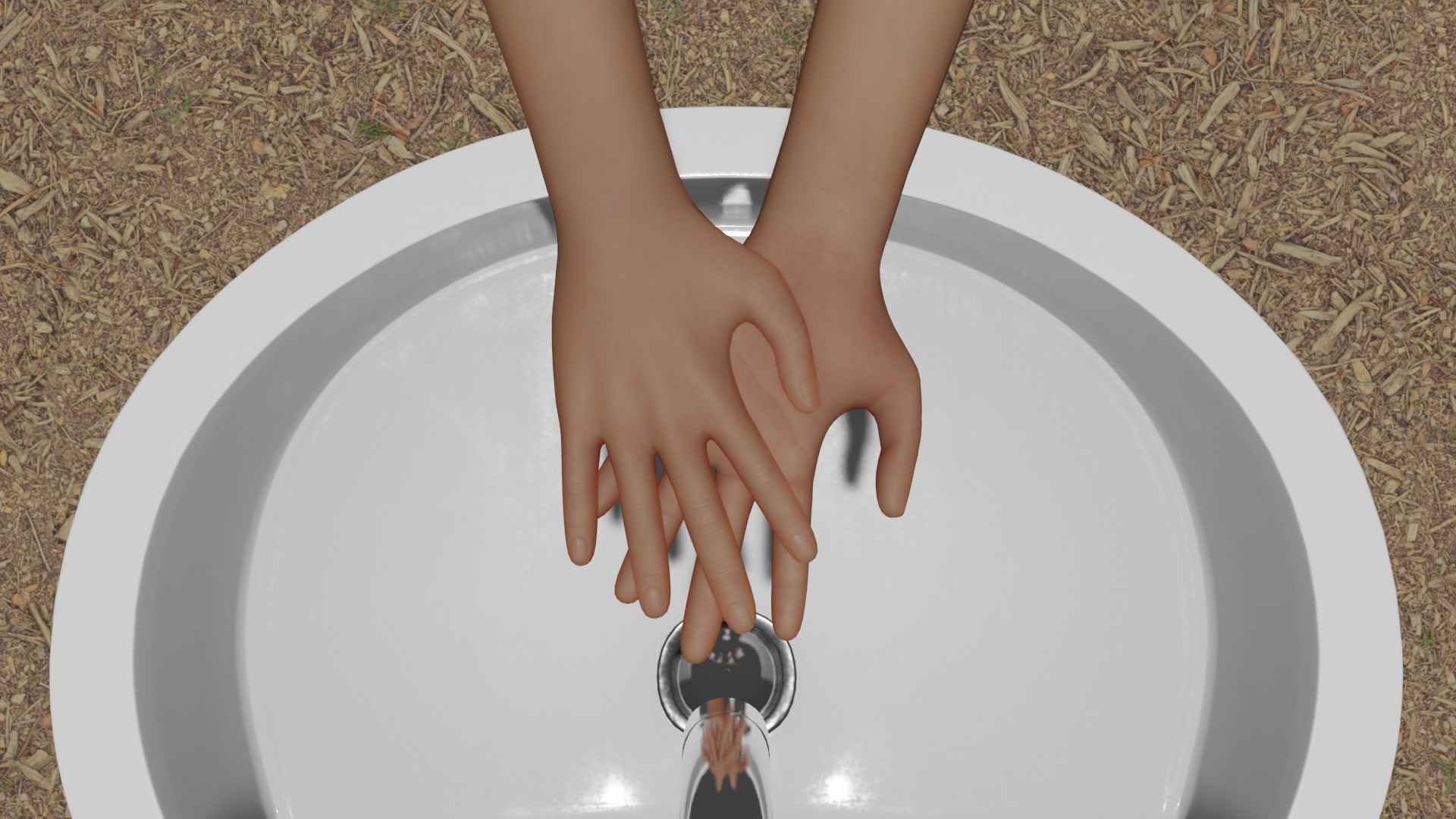}
\end{subfigure}%
\hspace{.1cm}
\begin{subfigure}[t]{0.22\linewidth}
    \includegraphics[width=\textwidth]{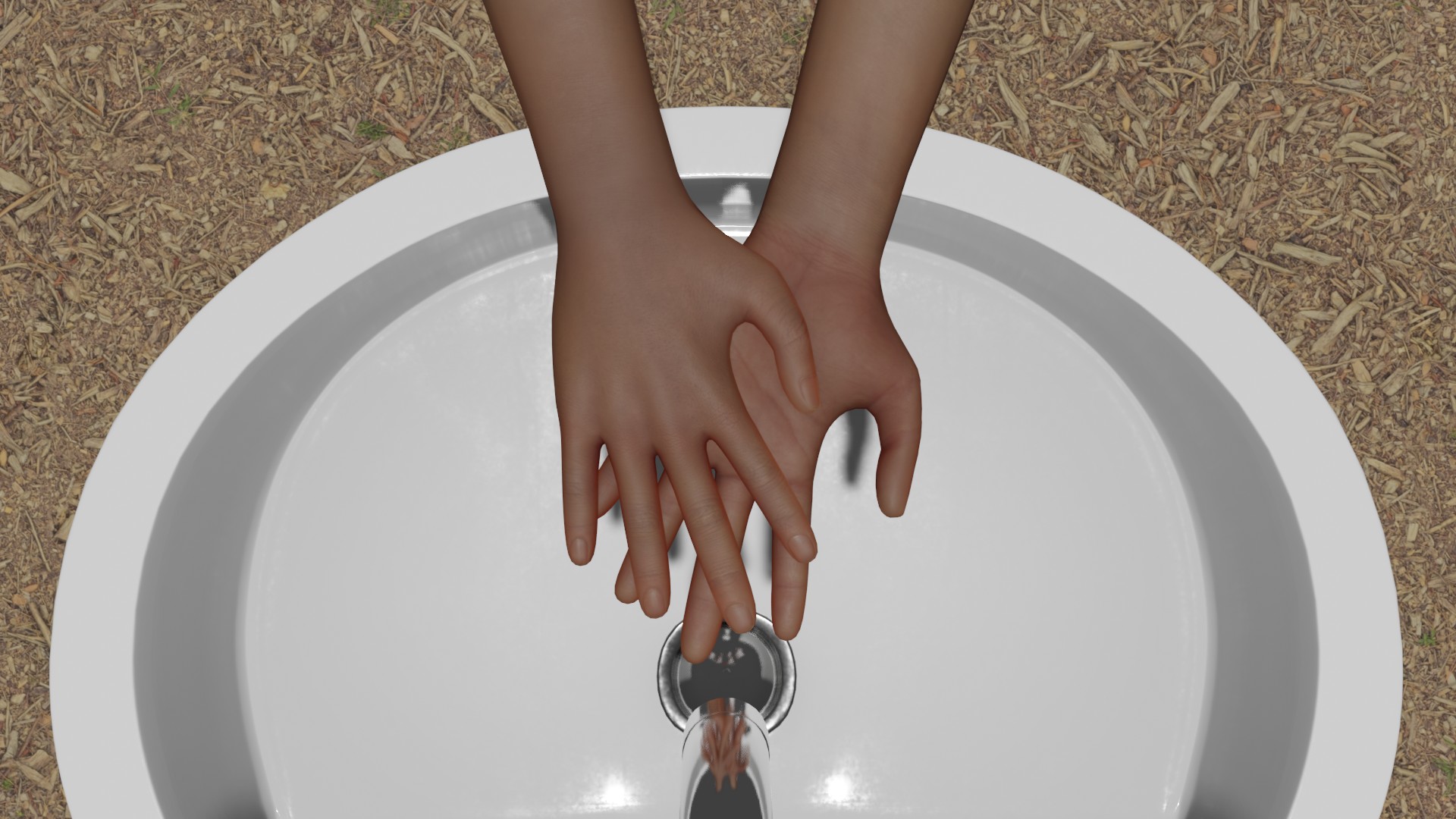}
\end{subfigure}%
\hspace{.1cm}
\begin{subfigure}[t]{0.22\linewidth}
    \includegraphics[width=\textwidth]{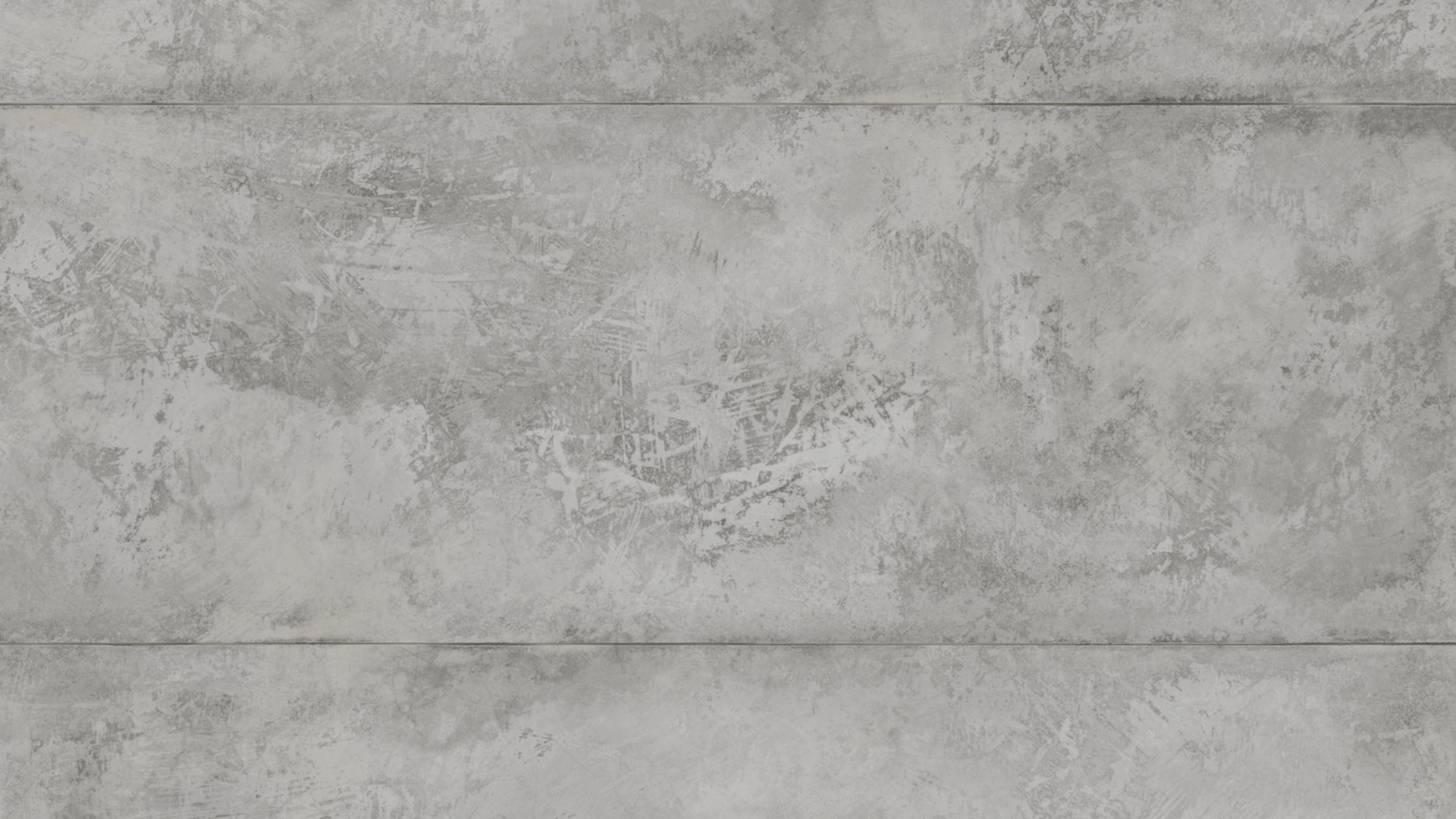}
\end{subfigure}%
\hspace{.1cm}
\begin{subfigure}[t]{0.22\linewidth}
    \includegraphics[width=\textwidth]{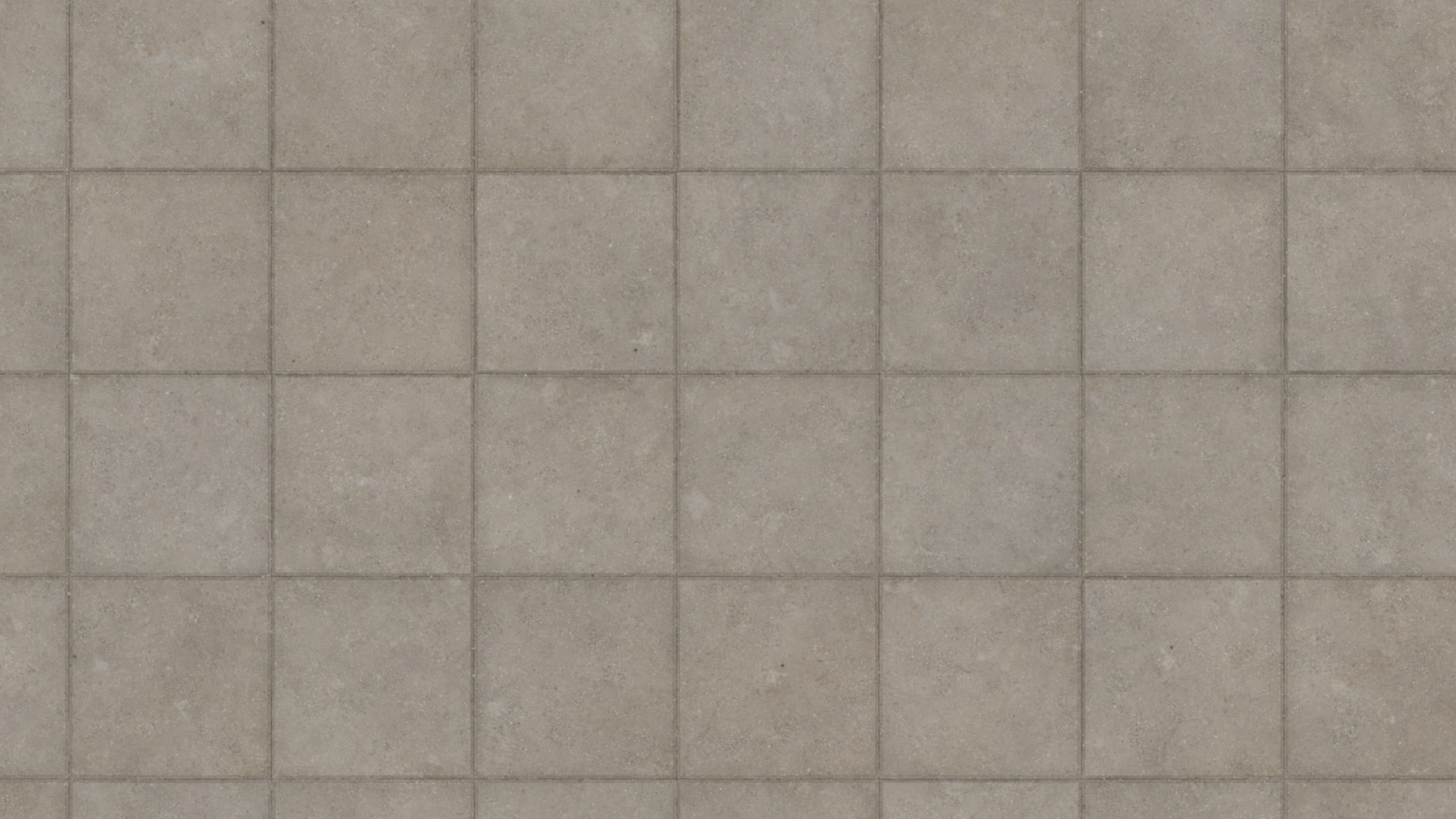}
\end{subfigure}
\vspace{.1cm}
\begin{subfigure}[t]{0.22\linewidth}
    \includegraphics[width=\textwidth]{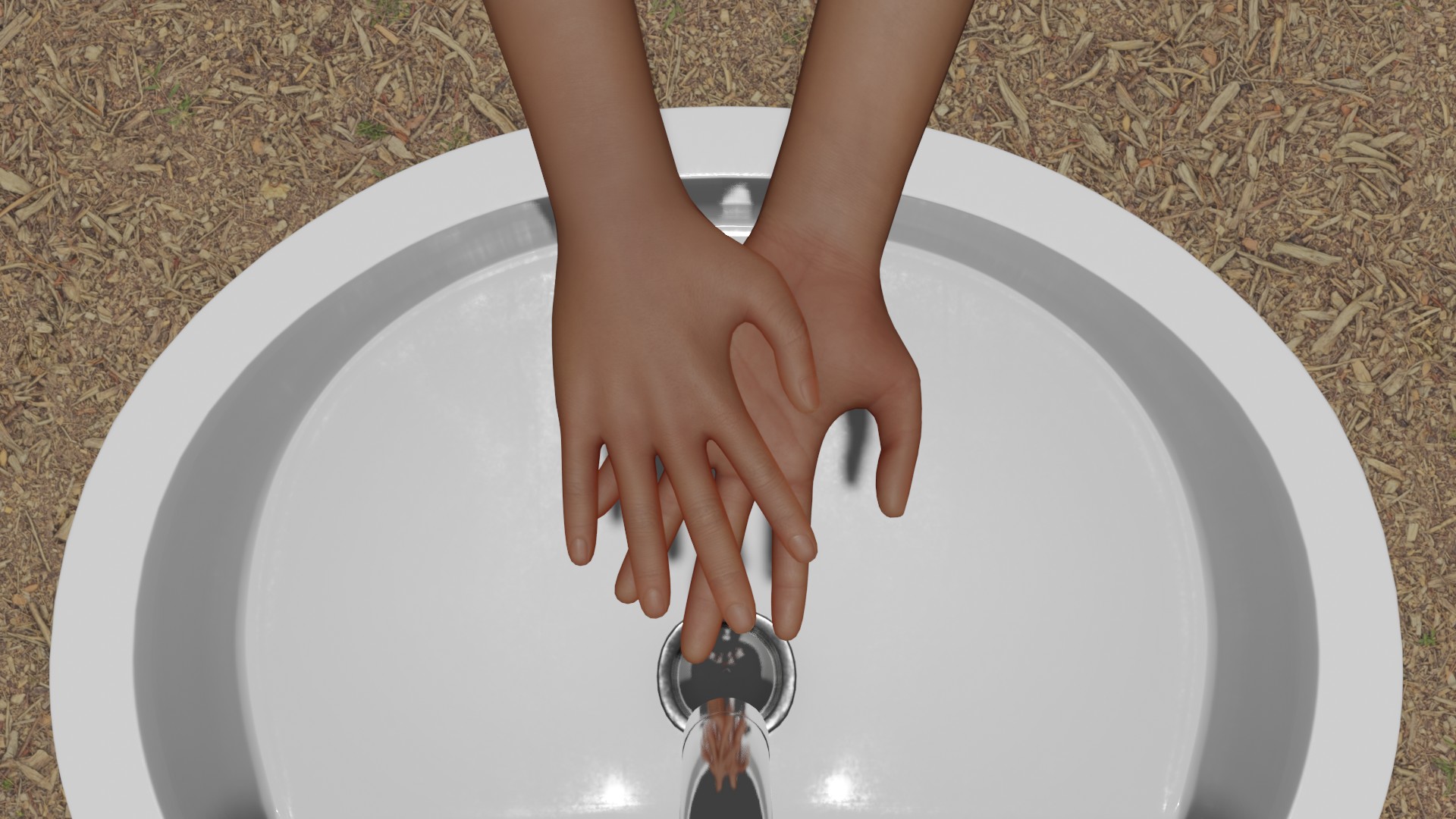}
\end{subfigure}%
\hspace{.1cm}
\begin{subfigure}[t]{0.22\linewidth}
    \includegraphics[width=\textwidth]{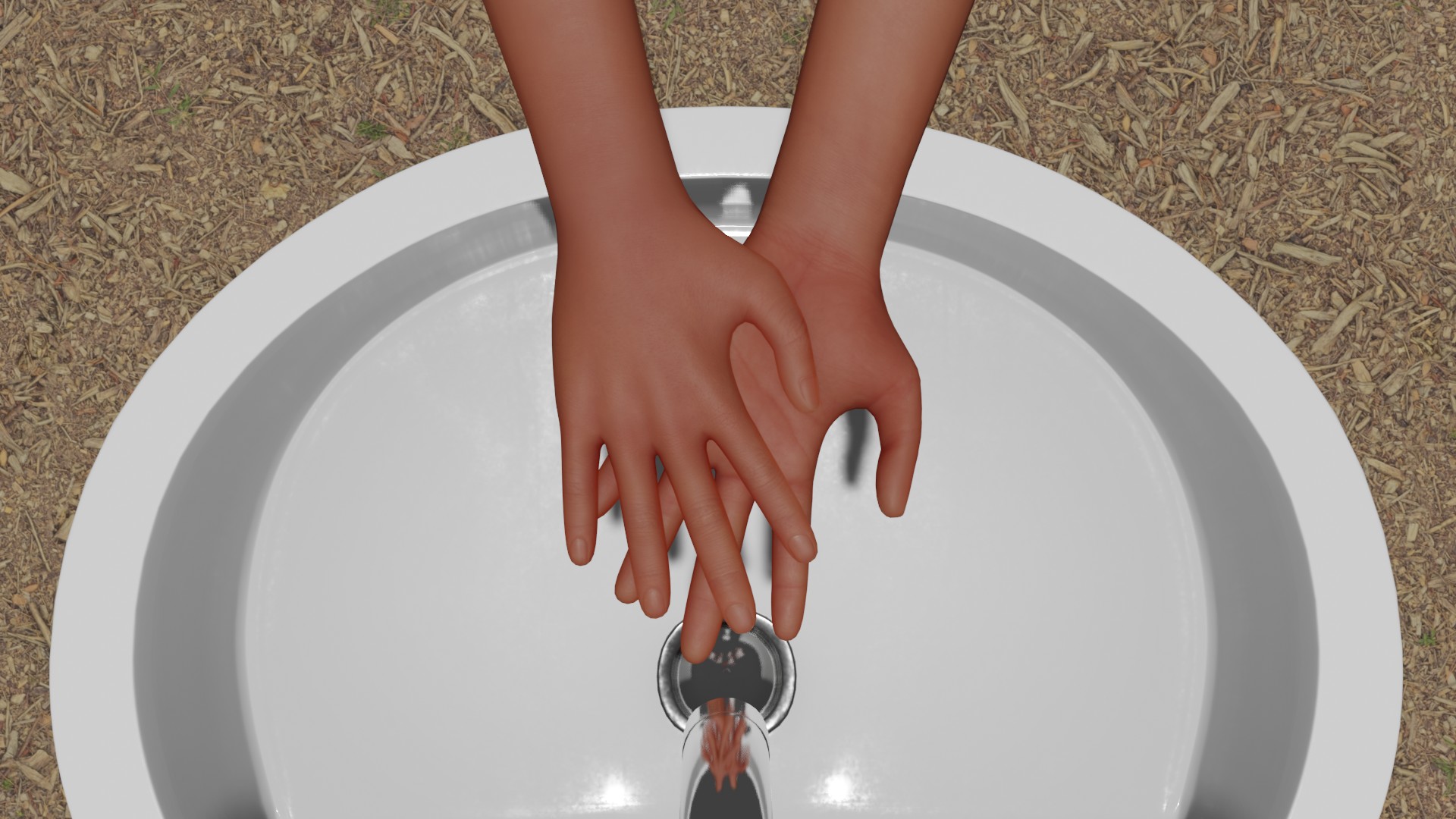}
\end{subfigure}%
\hspace{.1cm}
\begin{subfigure}[t]{0.22\linewidth}
    \includegraphics[width=\textwidth]{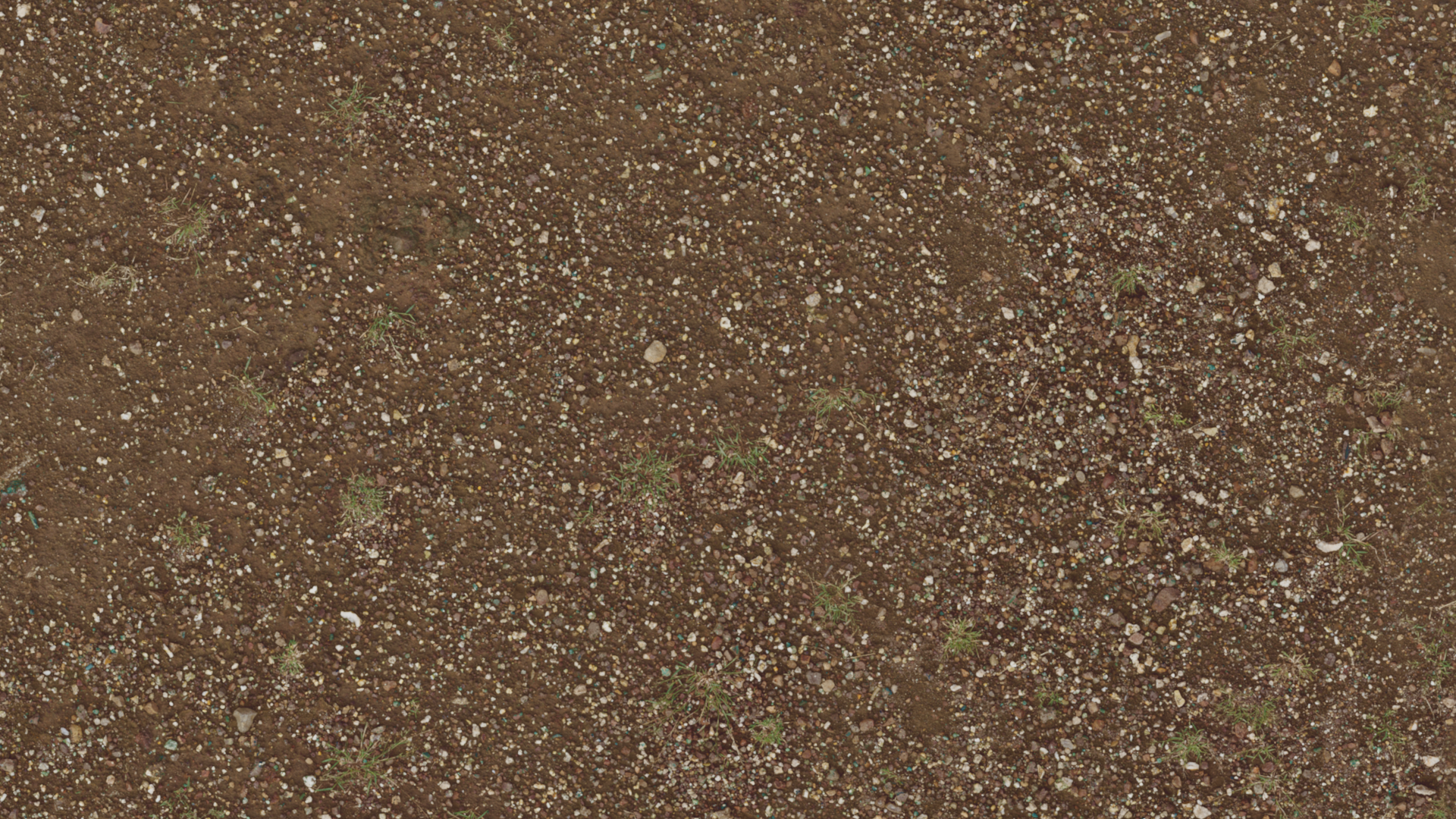}
\end{subfigure}%
\hspace{.1cm}
\begin{subfigure}[t]{0.22\linewidth}
    \includegraphics[width=\textwidth]{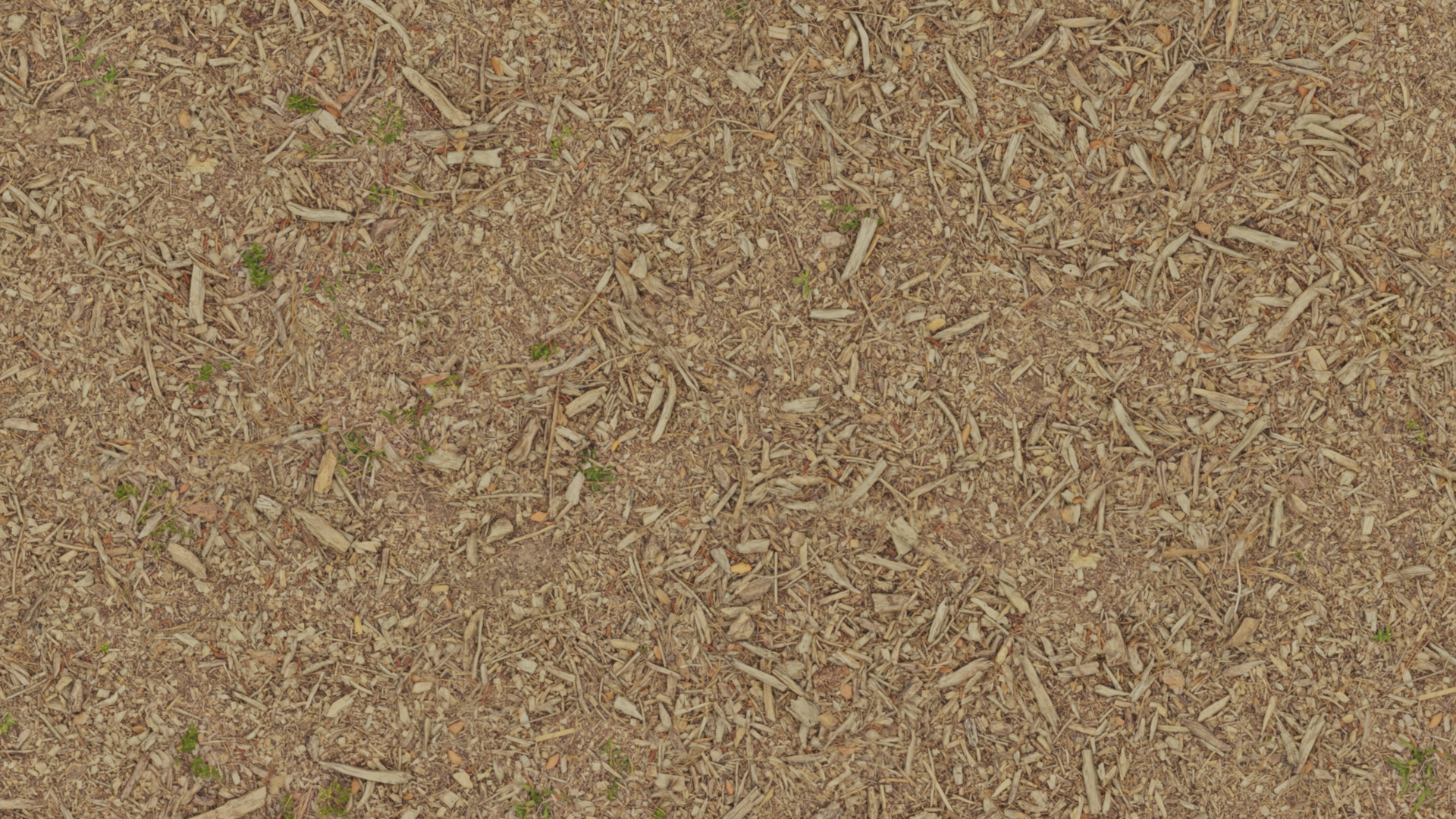}
\end{subfigure}
\vspace{.1cm}
\begin{subfigure}[t]{0.22\linewidth}
    \includegraphics[width=\linewidth]{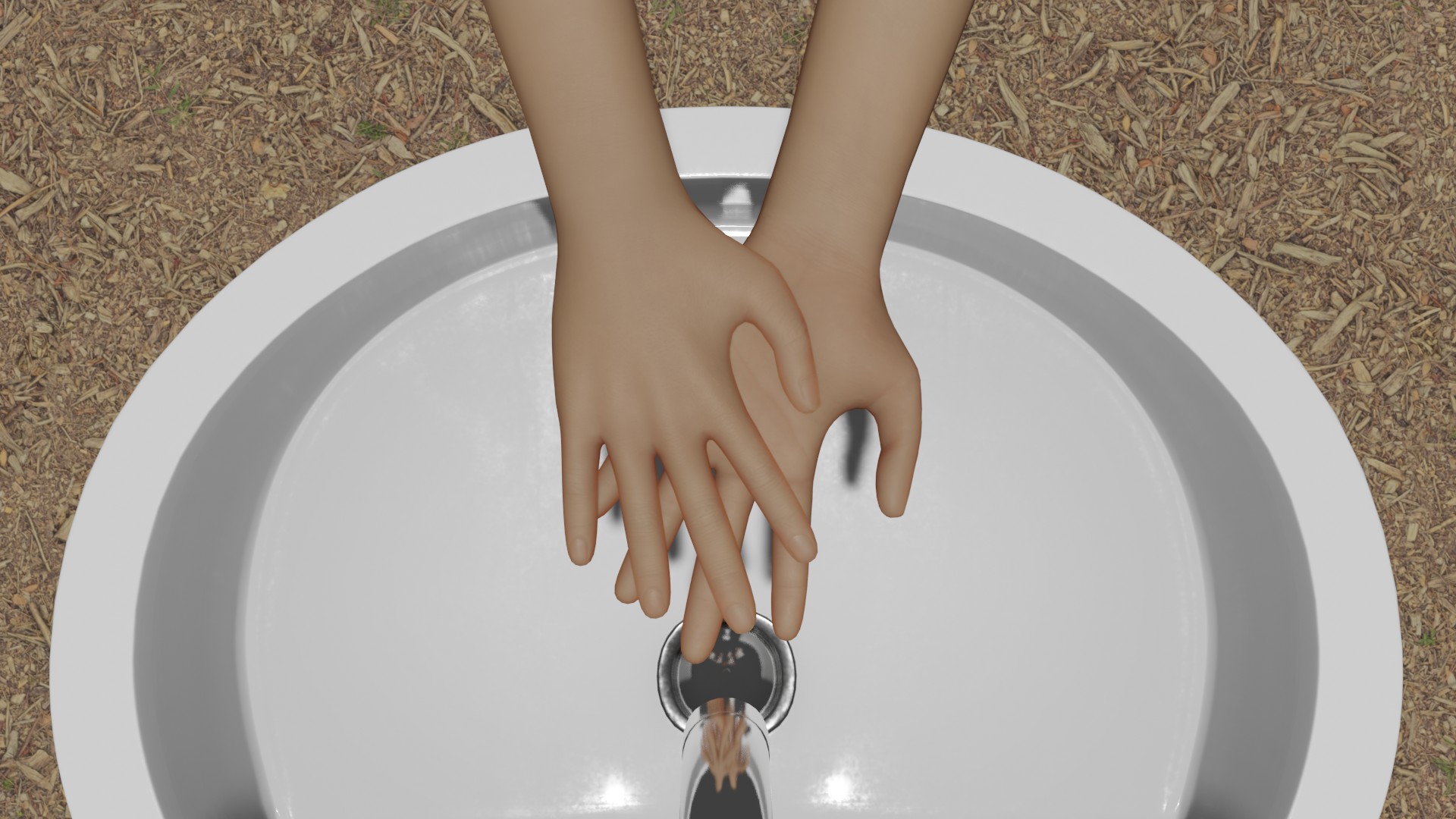}
\end{subfigure}%
\hspace{.1cm}
\begin{subfigure}[t]{0.22\linewidth}
    \includegraphics[width=\textwidth]{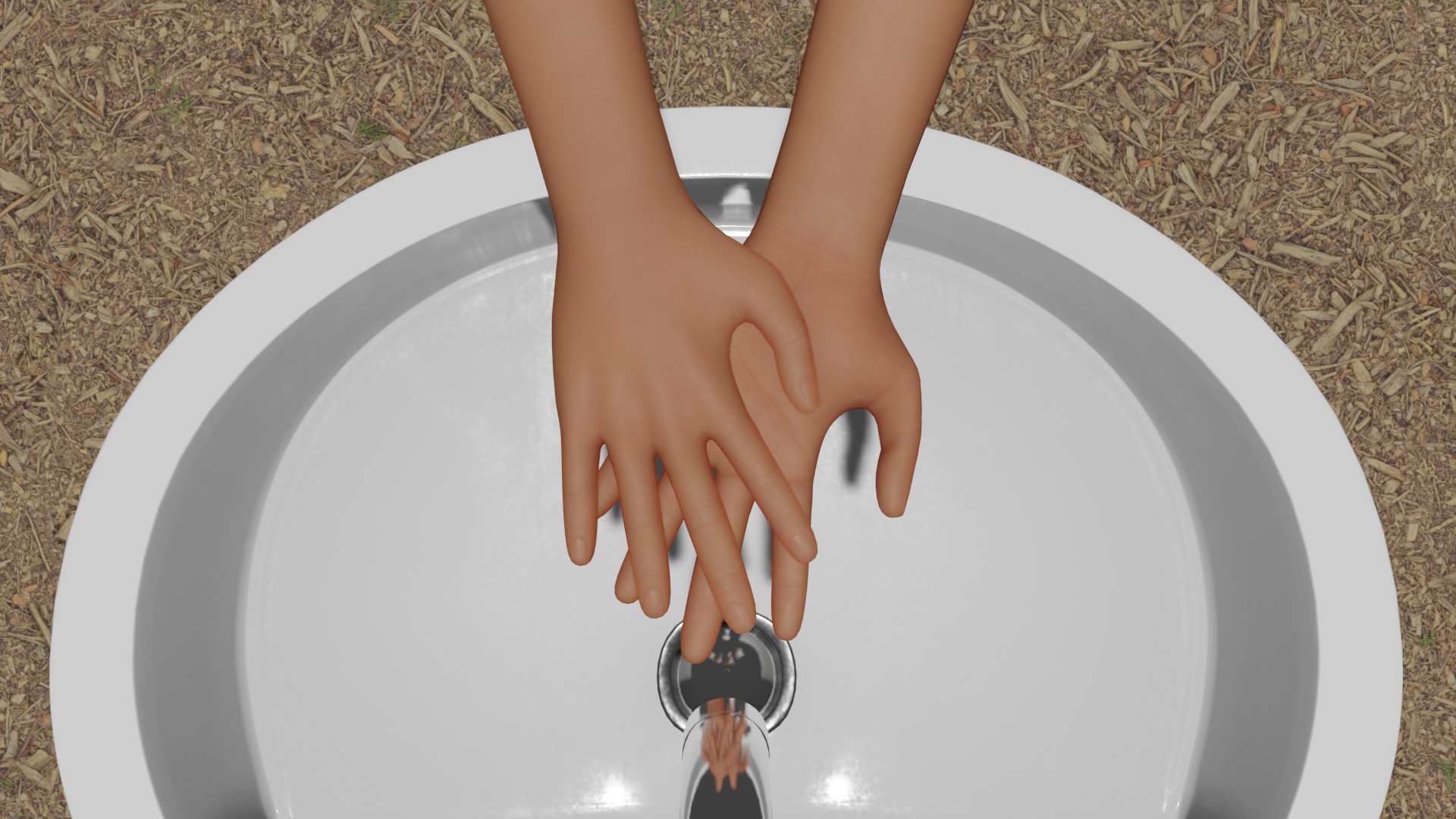}
\end{subfigure}%
\hspace{.1cm}
\begin{subfigure}[t]{0.22\linewidth}
    \includegraphics[width=\textwidth]{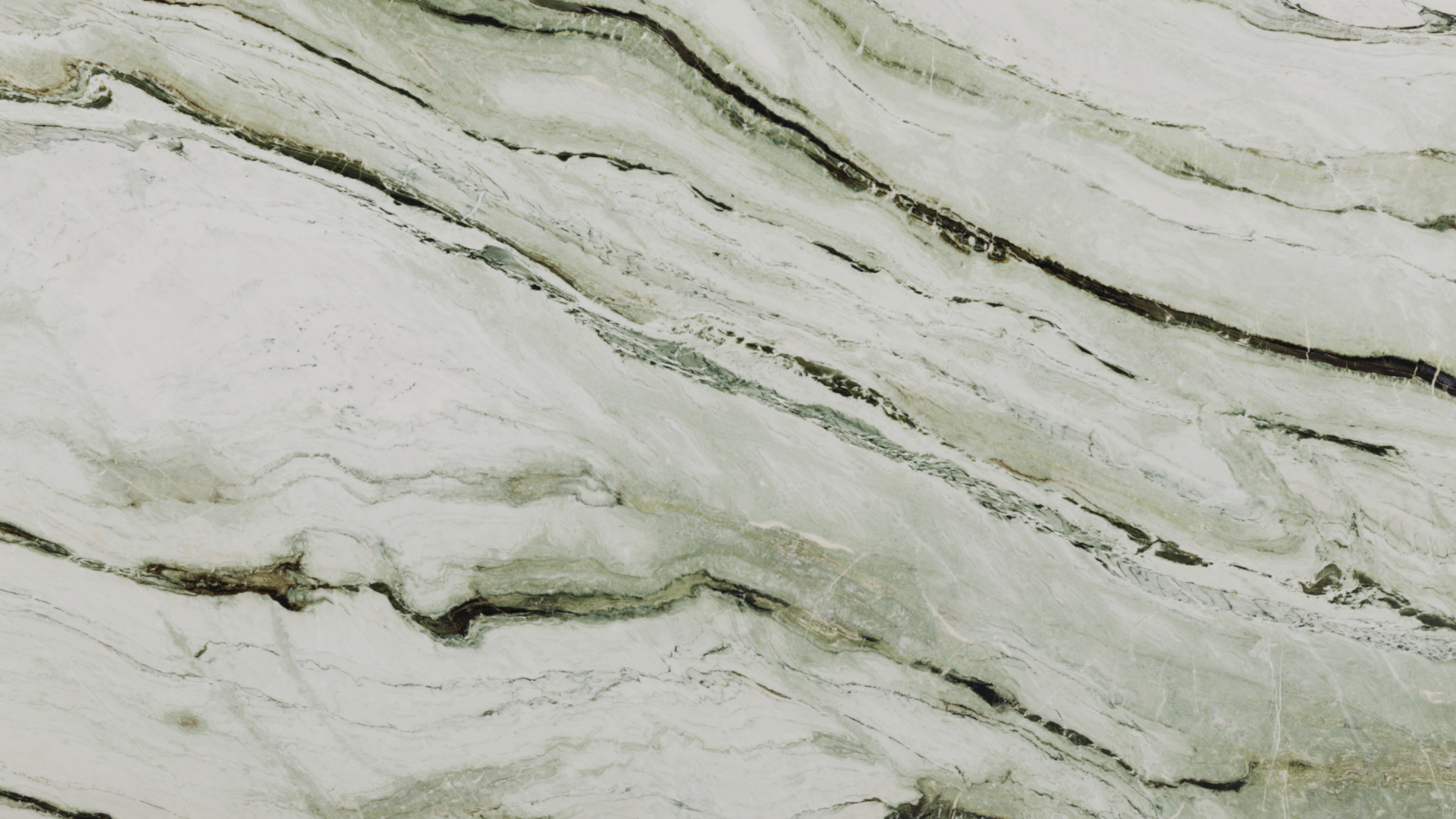}
\end{subfigure}%
\hspace{.1cm}
\begin{subfigure}[t]{0.22\linewidth}
    \includegraphics[width=\textwidth]{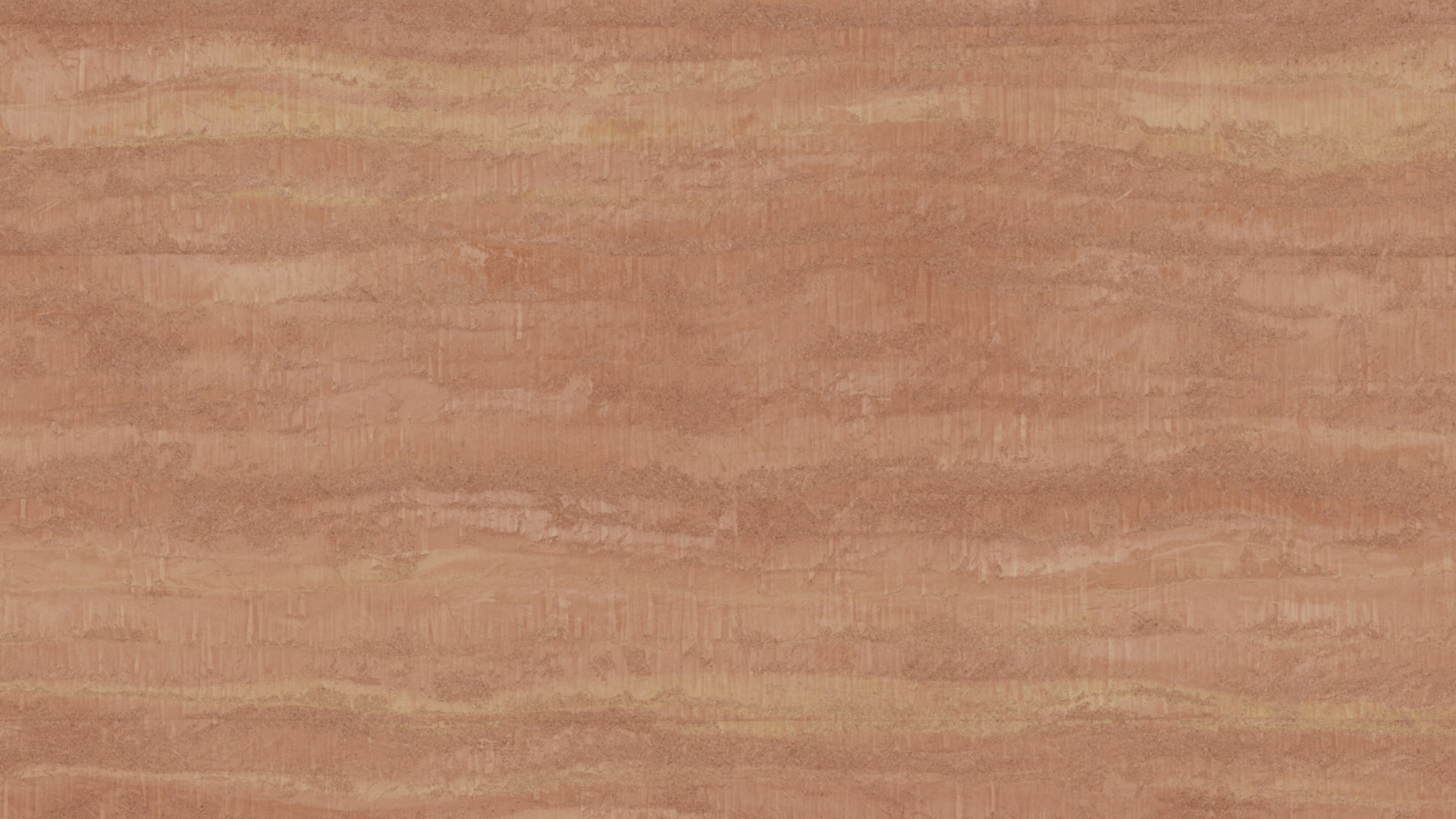}
\end{subfigure}
\vspace{.1cm}
\begin{subfigure}[t]{0.22\linewidth}
    \includegraphics[width=\textwidth]{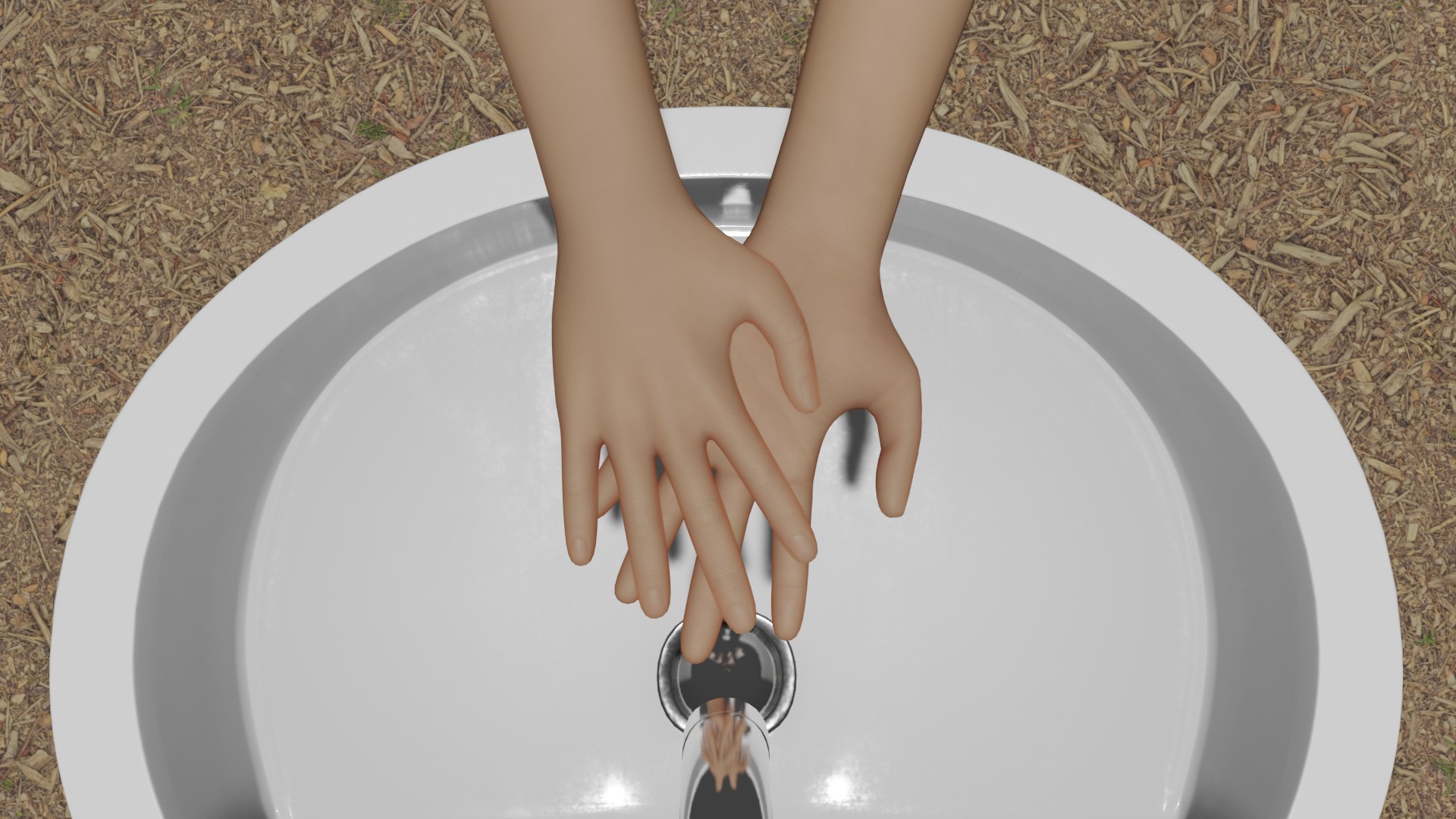}
\end{subfigure}%
\hspace{.1cm}
\begin{subfigure}[t]{0.22\linewidth}
    \includegraphics[width=\textwidth]{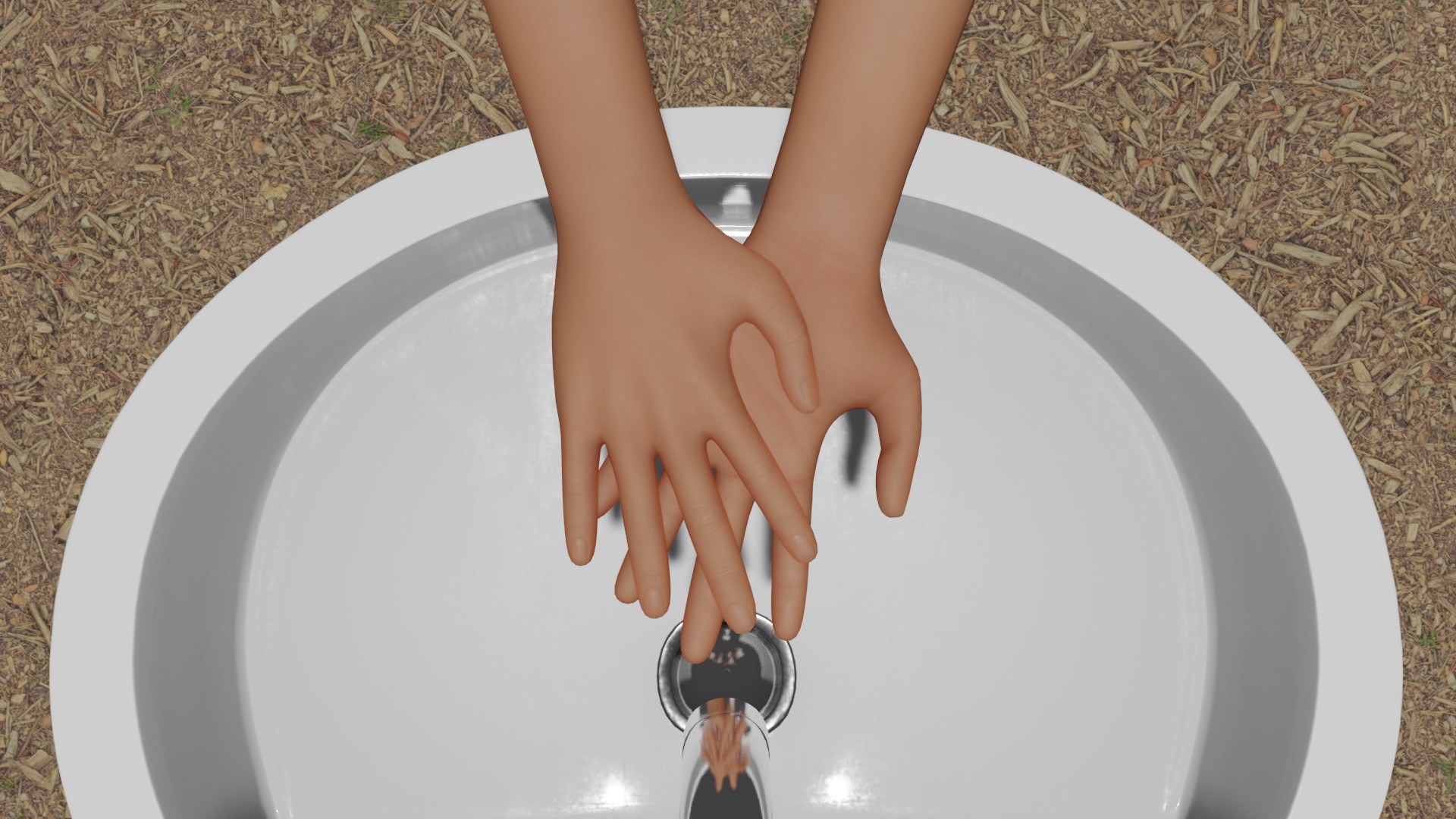}
\end{subfigure}%
\hspace{.1cm}
\begin{subfigure}[t]{0.22\linewidth}
    \includegraphics[width=\textwidth]{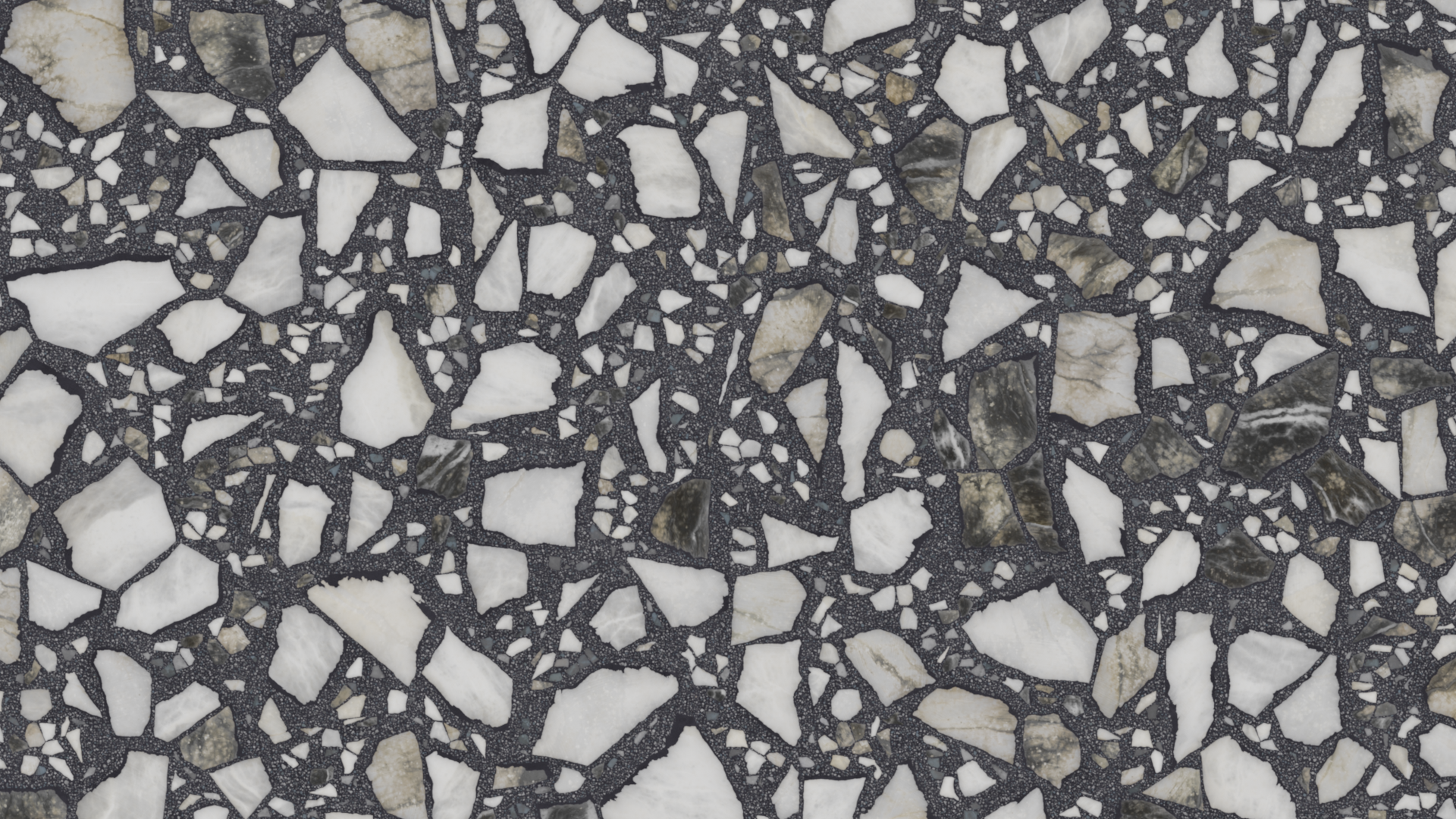}
\end{subfigure}%
\hspace{.1cm}
\begin{subfigure}[t]{0.22\linewidth}
    \includegraphics[width=\textwidth]{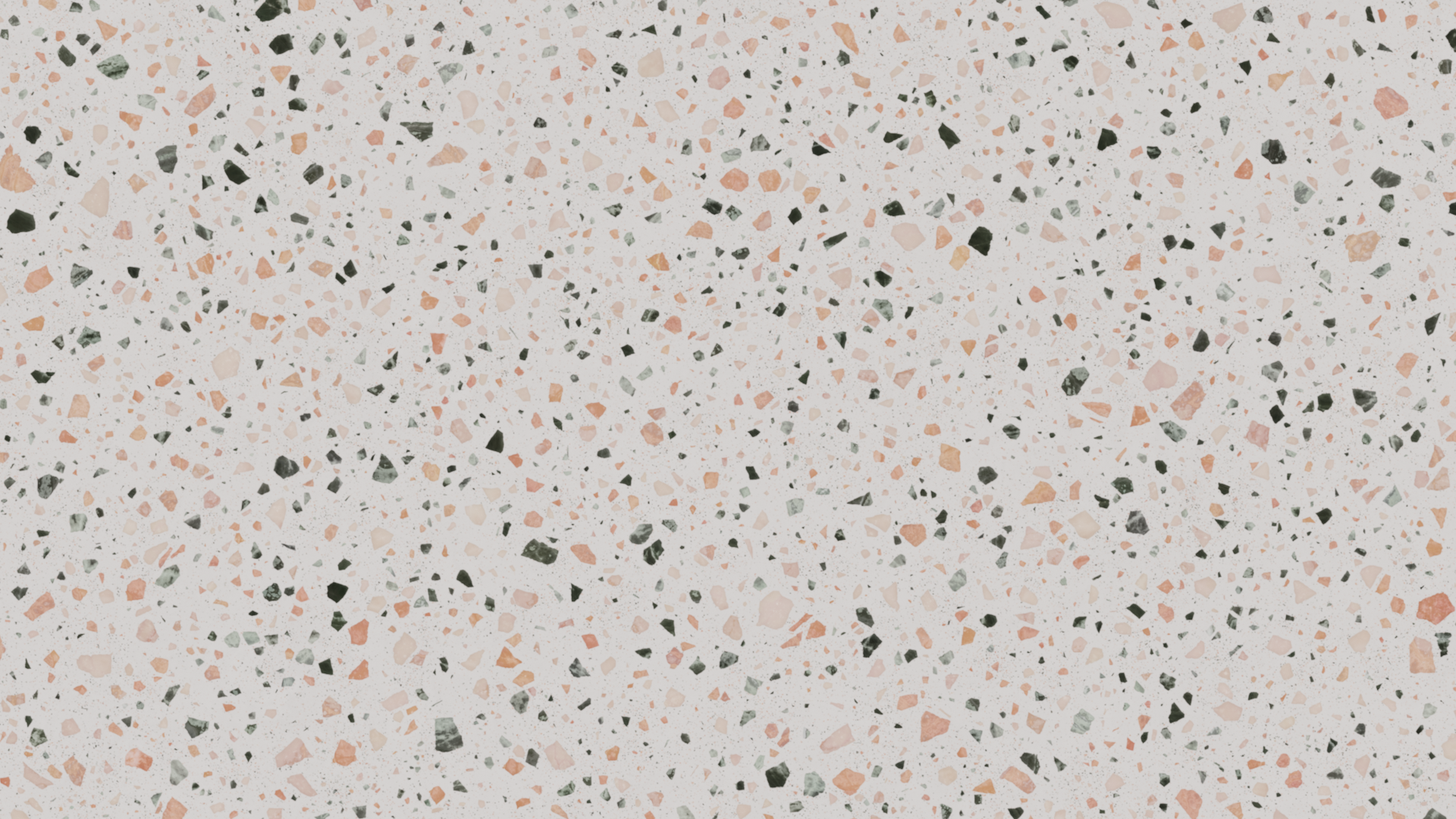}
\end{subfigure}
\vspace{.1cm}
\begin{subfigure}[t]{0.22\linewidth}
    \includegraphics[width=\textwidth]{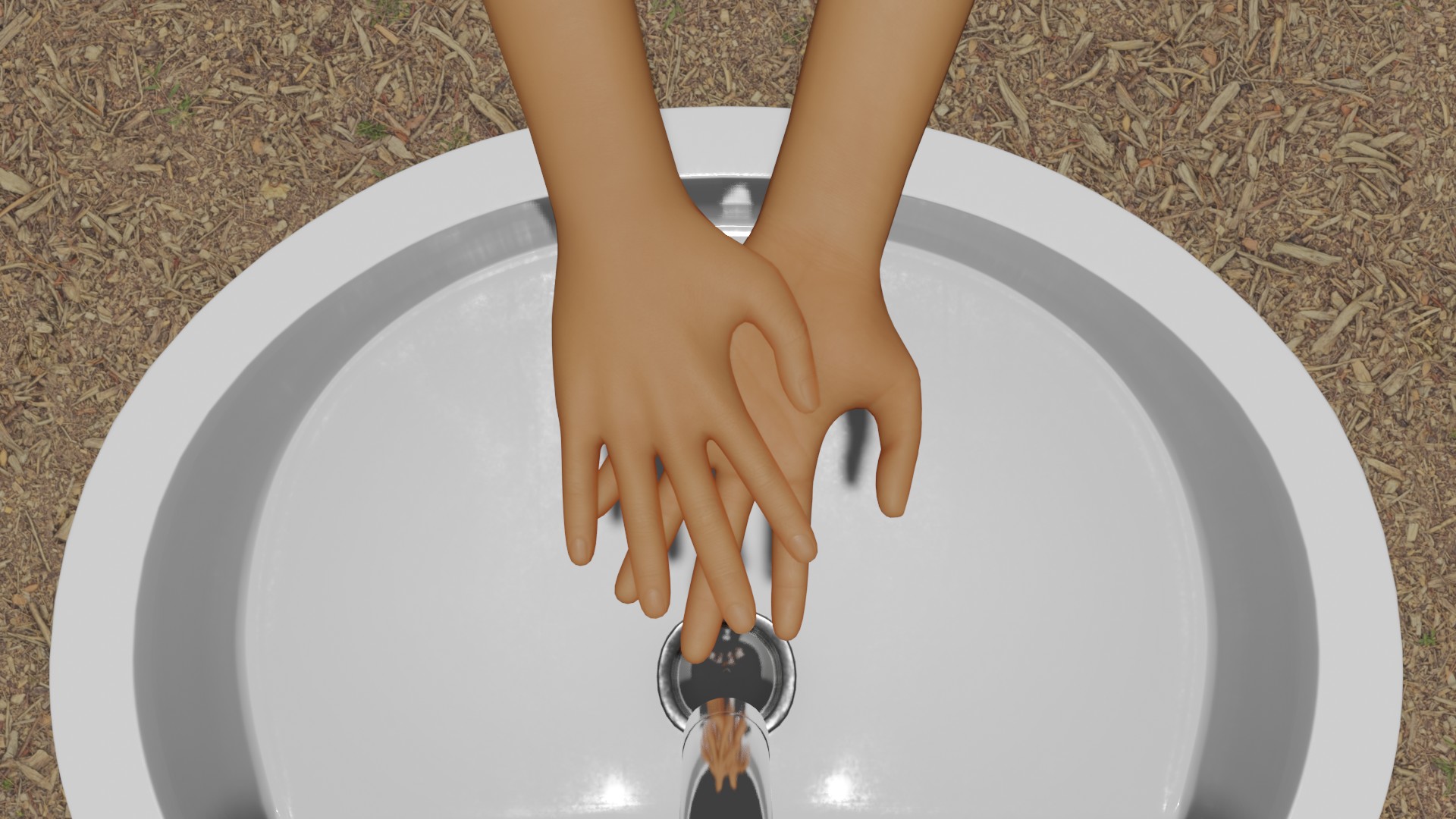}
\end{subfigure}%
\hspace{.1cm}
\begin{subfigure}[t]{0.22\linewidth}
    \includegraphics[width=\textwidth]{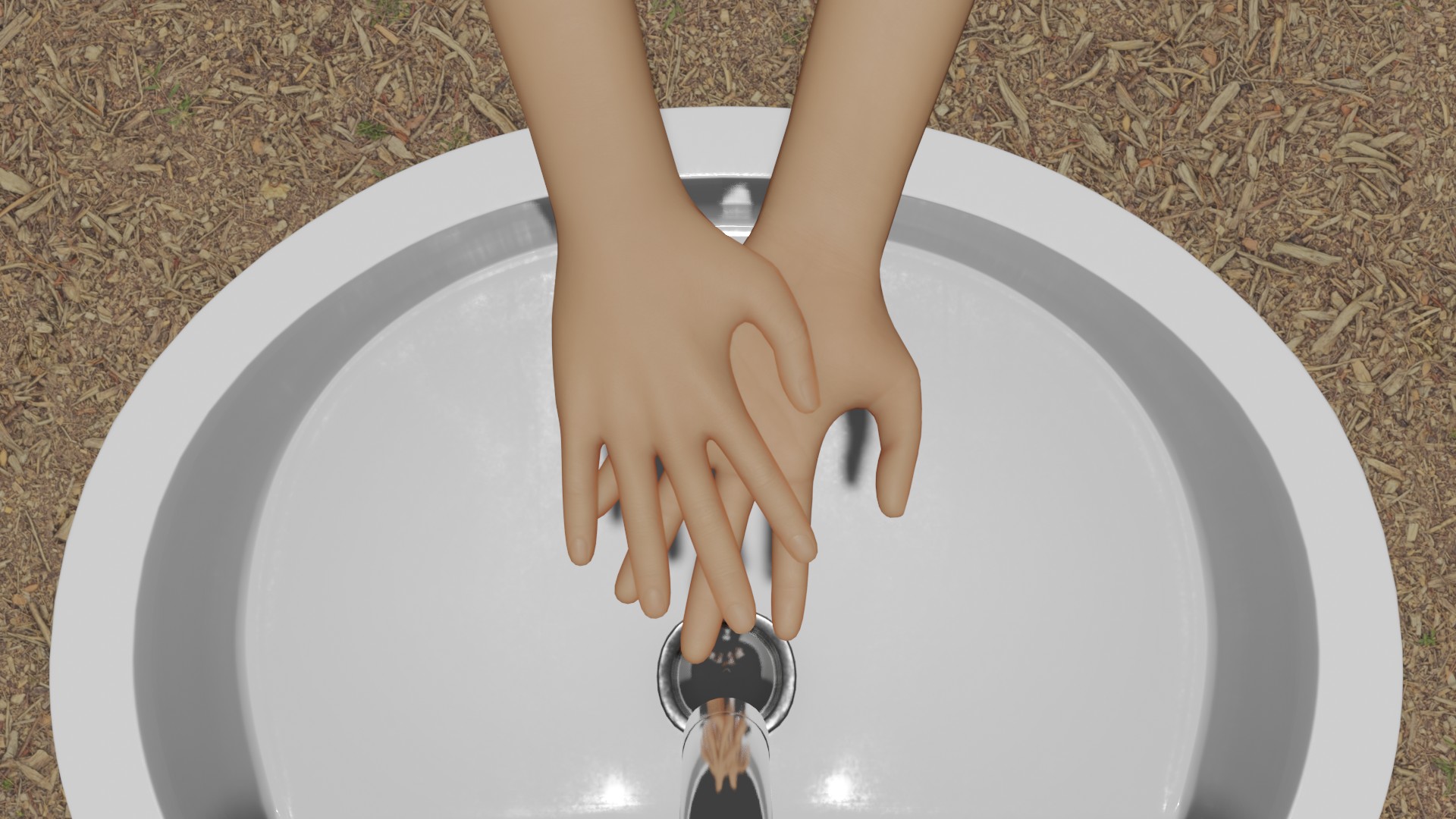}
\end{subfigure}%
\hspace{.1cm}
\begin{subfigure}[t]{0.22\linewidth}
    \includegraphics[width=\textwidth]{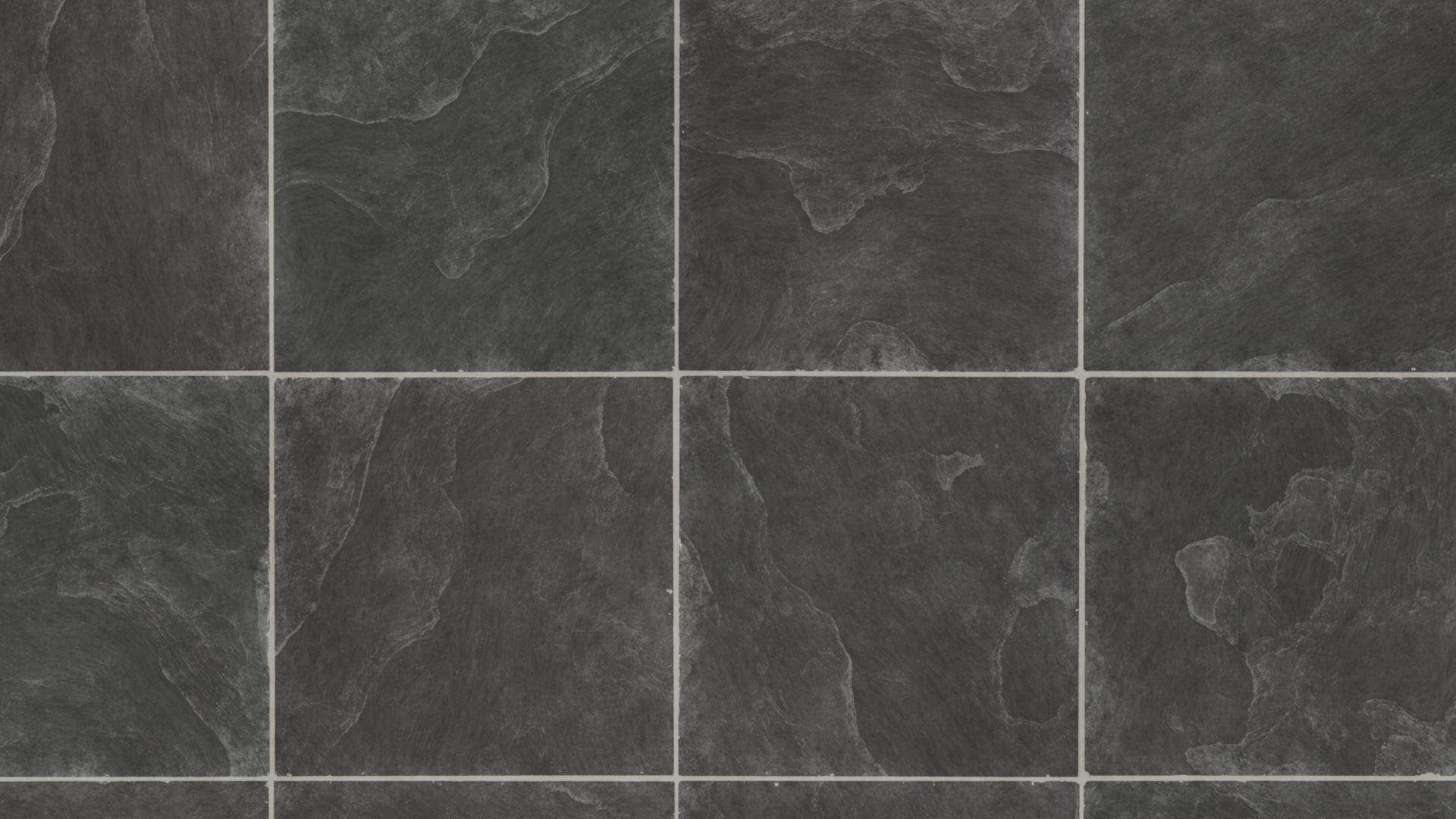}
\end{subfigure}%
\hspace{.1cm}
\begin{subfigure}[t]{0.22\linewidth}
    \includegraphics[width=\textwidth]{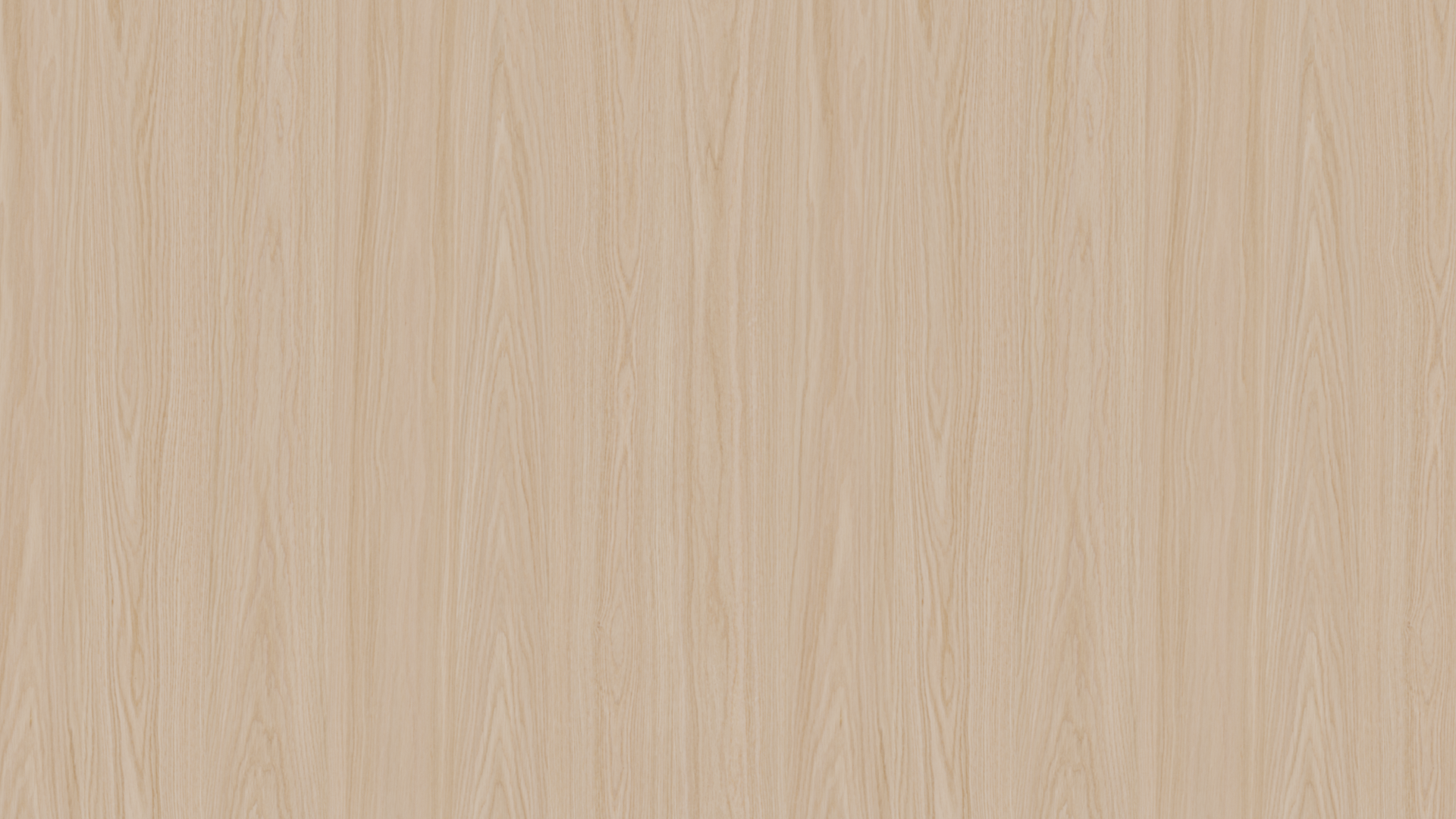}
\end{subfigure}%

\caption{Skin tones and background textures.}
\label{fig:skin_and_background}
\end{figure}

To make our synthetic images reflect real-world diversity, we apply 10 different skin tones and 10 different background textures. The left two columns of Figure~\ref{fig:skin_and_background} illustrate the 10 different skin tones we apply and the right two columns show the 10 different background textures. 
The backgrounds include both common indoor and outdoor floor textures to make the synthetic images look more realistic. 
Since the primary application setting of handwashing recognition will be outdoor, we use the sun as the light source in Blender. 

\begin{figure}[htb!]
\centering

\begin{subfigure}[t]{0.47\linewidth}
    \includegraphics[width=\textwidth]{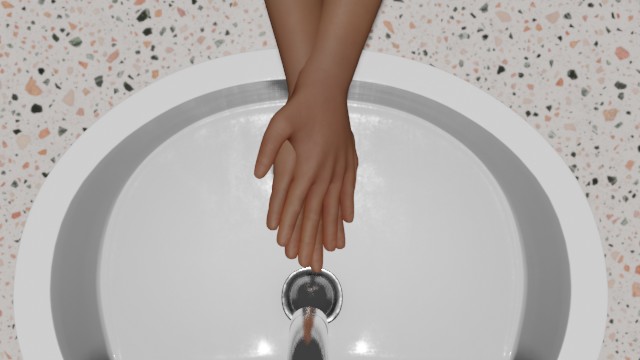}
    \caption{Rub back $-45^{\circ}$.}
\end{subfigure}%
\hspace{.1cm}
\begin{subfigure}[t]{0.47\linewidth}
    \includegraphics[width=\textwidth]{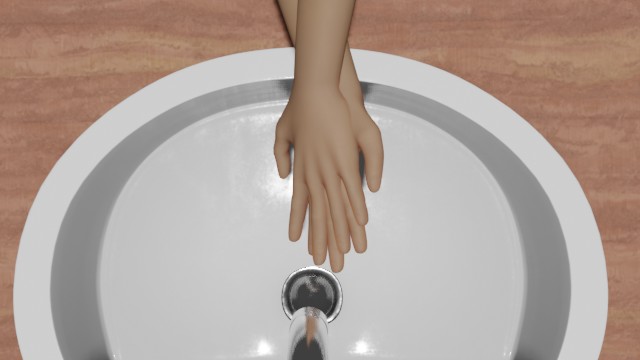}
    \caption{Rub back $45^{\circ}$.}
\end{subfigure}
\vspace{.1cm}
\begin{subfigure}[t]{0.47\linewidth}
    \includegraphics[width=\textwidth]{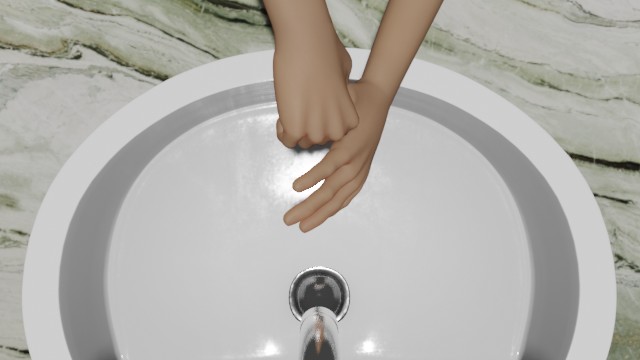}
    \caption{Rub thumb $-45^{\circ}$.}
\end{subfigure}%
\hspace{.1cm}
\begin{subfigure}[t]{0.47\linewidth}
    \includegraphics[width=\textwidth]{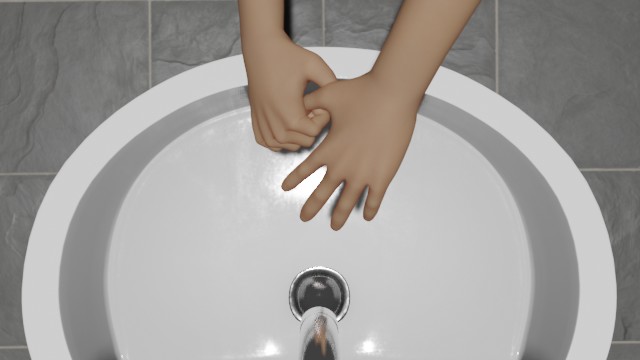}
    \caption{Rub thumb $45^{\circ}$.}
\end{subfigure}%

\caption{Examples of different hand poses.}
\label{fig:rotated_hands_ex}
\end{figure}

Most importantly, we vary hand poses and introduce controlled distribution shift by rotating the hands. Hand poses are quantified by the degree of rotation. We define the \textbf{canonical hand poses} as the 0-degree poses, as shown in Figure~\ref{fig:action_ex}. We rotate the hands for all actions from $-90^{\circ}$ to $90^{\circ}$ with respect to their individual 0-degree poses. Figure~\ref{fig:rotated_hands_ex} shows examples of rotated hand poses. The range for hand rotation angles is chosen based on realistic and possible non-canonical hand poses that real humans can perform. 

In total, we generate 518,000 images for hand poses with varied background and skin tones. 

\subsection{Synthetic Data Creation (Shadow)}
To study the impact of shadow, we generate images with different properties of shadow including size, intensity, and placement. We create shadow in images by placing a cylindrical pole object between the light source and the hands. The properties of the pole are fully adjustable so we can synthesize images with different shadows. Again due to the computational constraints and the amount of images, we only render shadow images for two actions: rub back and rub thumb. In addition, only one of the dual poses is considered for these two actions. 

\begin{figure}[htb!]
\centering

\begin{subfigure}[t]{0.31\linewidth}
    \includegraphics[width=\textwidth]{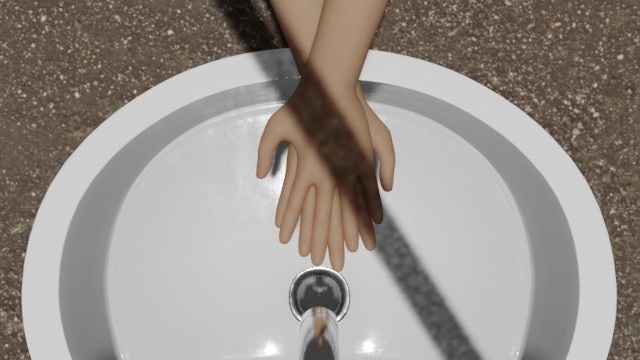}
    \caption{Shadow size 1.}
\end{subfigure}%
\hspace{.1cm}
\begin{subfigure}[t]{0.31\linewidth}
    \includegraphics[width=\textwidth]{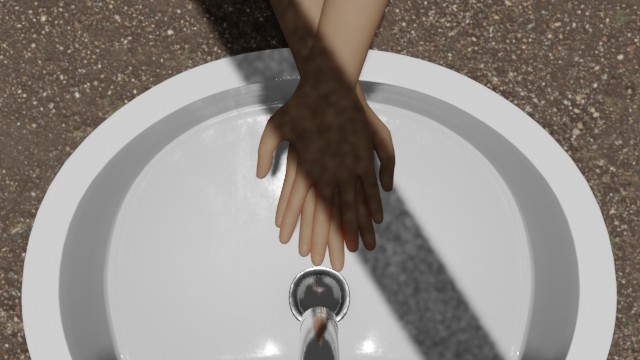}
    \caption{Shadow size 2.}
\end{subfigure}%
\hspace{.1cm}
\begin{subfigure}[t]{0.31\linewidth}
    \includegraphics[width=\textwidth]{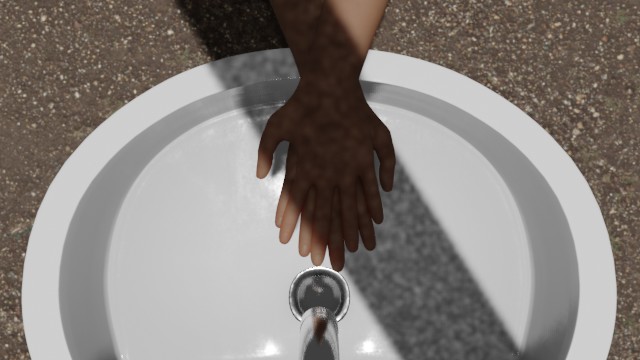}
    \caption{Shadow size 3.}
\end{subfigure}%

\caption{Examples of different shadow size.}
\label{fig:shadow_pole}
\end{figure}

First, we synthesize shadow images to study the impact of shadow size. Shadow size can be controlled by adjusting the width of the cylindrical pole. Figure~\ref{fig:shadow_pole} shows an example of 3 different pole widths that correspond to 3 different shadow sizes.

\begin{figure}[htb!]
\centering

\begin{subfigure}[t]{0.22\linewidth}
    \includegraphics[width=\textwidth]{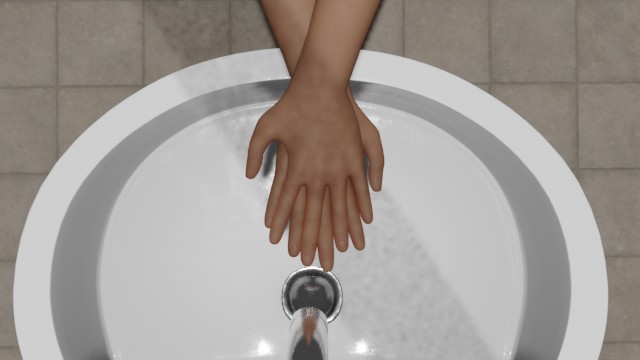}
    \caption{Alpha=0.2.}
\end{subfigure}%
\hspace{.1cm}
\begin{subfigure}[t]{0.22\linewidth}
    \includegraphics[width=\textwidth]{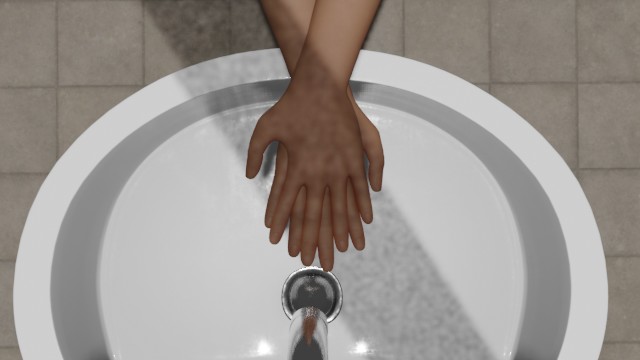}
    \caption{Alpha=0.4.}
\end{subfigure}%
\hspace{.1cm}
\begin{subfigure}[t]{0.22\linewidth}
    \includegraphics[width=\textwidth]{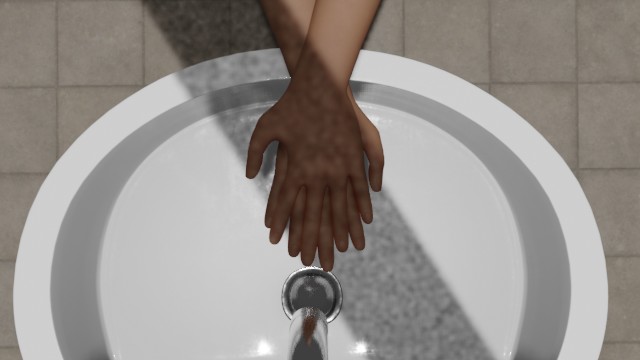}
    \caption{Alpha=0.6.}
\end{subfigure}%
\hspace{.2cm}
\begin{subfigure}[t]{0.22\linewidth}
    \includegraphics[width=\textwidth]{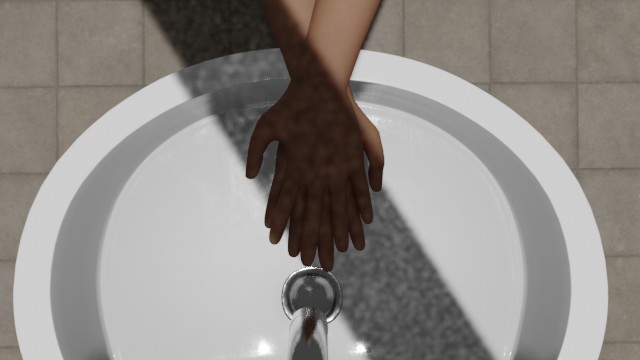}
    \caption{Alpha=0.8.}
\end{subfigure}%

\caption{Examples of different shadow intensity.}
\label{fig:shadow_alpha}
\end{figure}

Then, we synthesize shadow images to study the impact of shadow intensity. In our data, shadow intensity is represented by the darkness of the shadow. To adjust the darkness of shadow, we add translucency to the cylindrical pole and adjust the value of the alpha channel to change the amount of light that passes through. Figure~\ref{fig:shadow_alpha} shows sample images of the four different shadow intensities we apply. Lower alpha values correspond to lighter shadows, and vice versa. 

\begin{figure}[htb!]
\centering

\begin{subfigure}[t]{0.31\linewidth}
    \includegraphics[width=\textwidth]{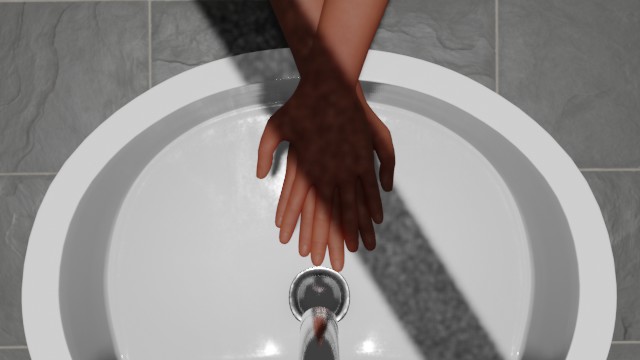}
    \caption{Transl. 1 rot. 1.}
    \label{subfig: loc1rot1}
\end{subfigure}%
\hspace{.1cm}
\begin{subfigure}[t]{0.31\linewidth}
    \includegraphics[width=\textwidth]{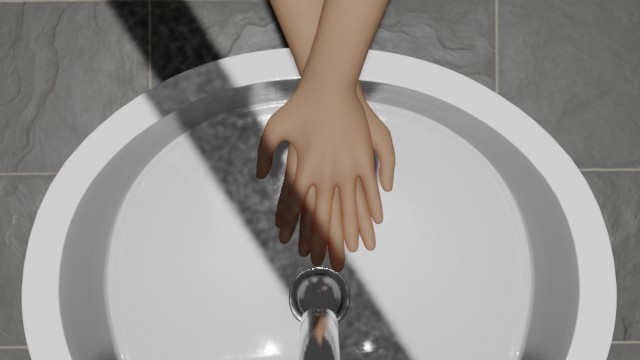}
    \caption{Transl. 2 rot. 1.}
    \label{subfig: loc2rot1}
\end{subfigure}%
\hspace{.1cm}
\begin{subfigure}[t]{0.31\linewidth}
    \includegraphics[width=\textwidth]{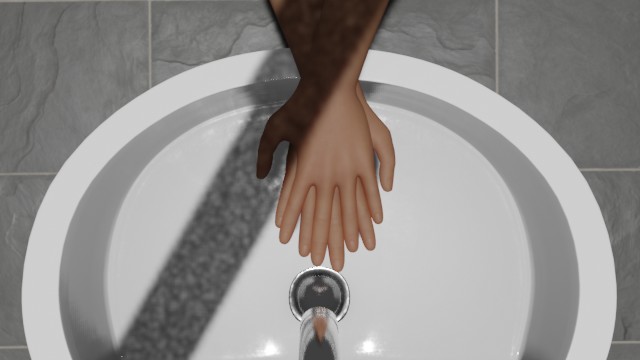}
    \caption{Transl. 1 rot. 2.}
    \label{loc1rot2}
\end{subfigure}%

\caption{Examples of different shadow placements.}
\label{fig:shadow_rot_loc}
\end{figure}

Next, we synthesize shadow images with different rotations and translations to study the impact of shadow placement. The motivation is that shadow can appear in different orientations and locations in outdoor conditions. Sometimes shadow can cover the entire palm or the back of hand, while other times it will only cover a portion of the fingertips. For a recognition system, it is critical for the system to see the regions where the two hands meet and interact. Thus, intuitively, shadow that covers more hand-interacting regions will cause more significant performance drop. We consider 2 different rotations and 2 different translations of the pole to simulate random placements of shadow, as shown in Figure~\ref{fig:shadow_rot_loc}.

In total, we generate 7.1M images for all shadow cases. 

\section{Experiments and Results}
\label{sec:exp_results}
In this section, we discuss details about our experiments and analyze results. We design 3 sets of experiments: (1) baseline experiments with non-shadow images to investigate the breakdown points caused by hand poses, (2) experiments using shadow images to study shadow's impact on overall performance and also breakdown points, and (3) experiments to explore the effectiveness of additional training poses as a mitigation strategy to pose-induced breakdown points. As mentioned in Section~\ref{subsec:classifier}, all experiments are conducted as image classification using MobileNetV3\cite{mbnetv3}.  

\subsection{Do hand poses impact performance?}
\label{subsec: baseline_exps}
\subsubsection{Experimental setup}
To study the impact of hand poses on the classifier, we consider our training data to consist only of hand images without shadow and in canonical poses ($0^{\circ}$). The five classes correspond to the 5 rubbing actions we defined earlier. 
The 0-degree and shadow-free data for all actions are split into training and testing data using a 1:1 ratio. We also ensure a balanced training set by including the same number of images for each action. Data from all other poses/angles are used as testing data. We fine-tune a pretrained MobileNetV3 network for 1,200 iterations using a batch size of 128. 

\subsubsection{Evaluation metrics}
We use top1 accuracy as the main evaluation metric and define the breakdown points with respect to hand poses, using changes in the top1 accuracy. The breakdown point is designed to measure the hand rotational angle that causes the classifier's performance to drop sharply or fall below a certain satisfactory threshold. First, we define two criteria: 
(1) the top1 accuracy drops below 60\%, and (2) the top1 accuracy drops by more than 15\% for a consecutive 5-degree hand rotation. Starting from the 0-degree pose, we find breakdown points for both positive and negative rotational angles. A breakdown point is found when either criteria is met for each direction of pose change. 

\subsubsection{Results and discussion}
\begin{figure}[t]
    \centering
    \includegraphics[width=0.9\linewidth]{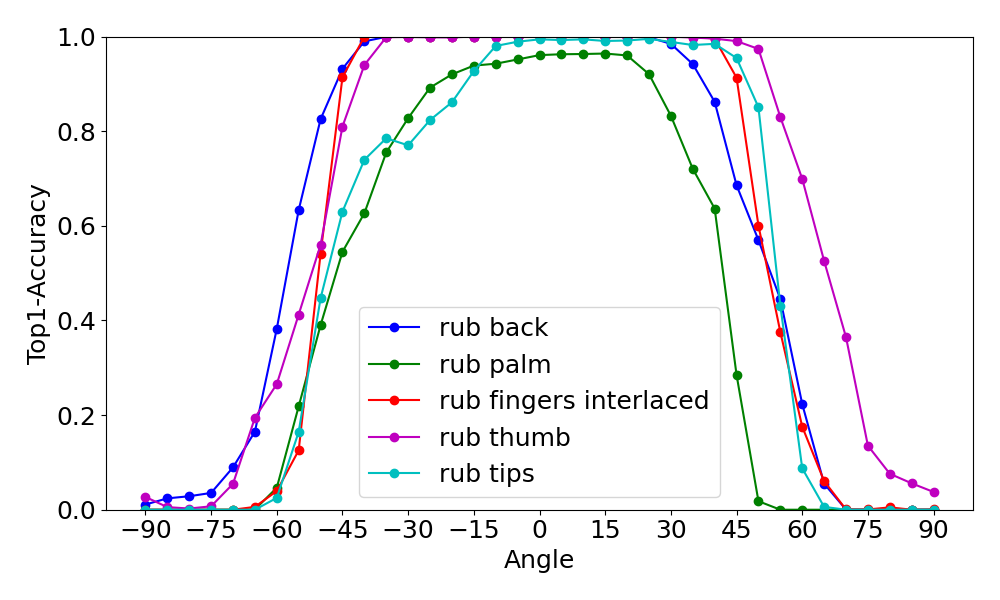}
    \caption{Top1 accuracy at different angles.}
    \label{fig:baseline_top1}
\end{figure}

\begin{figure}[htb!]
    \centering
    \includegraphics[width=0.85\linewidth]{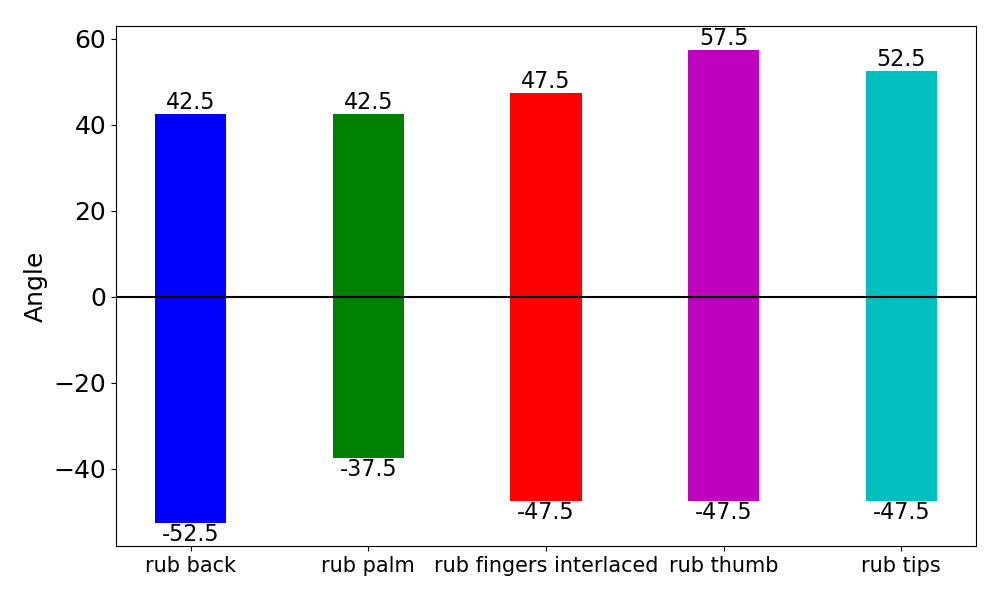}
    \caption{Breakdown points for 5 rubbing actions.}
    \label{fig:baseline_bdpts}
\end{figure}

Figure~\ref{fig:baseline_top1} shows the overall top1 accuracy for the 5 actions at different hand angles. Visually, it is clear that system performance remains high for small deviations in pose, but then drops significantly when the hand angle increases; this indicates the presence of breakdown points. Also, each action exhibits a different breakdown point. Figure~\ref{fig:baseline_bdpts} summarizes the breakdown points from both the positive angles and negative angles. By averaging the two breakdown points, we see that the breakdown points of rub thumb and rub tips occur slightly later than the other actions. This could be due to the uniqueness of hand poses associated with these two actions compared to other actions. More specifically, actions like grabbing the thumb or pinching the fingers together could be more distinguishable.

Overall, it is clear that hand poses do impact classification performance by degrading performance sharply after a certain angle, and this can be quantified by the breakdown points.
By using breakdown points computed with respect to hand pose, researchers can better understand when a system's performance starts to fall sharply and become unreliable. 

\subsection{Does shadow impact overall performance and the breakdown points?}
\label{subsec: shadow_exps}
\subsubsection{Experimental setup}
To measure the impact of shadow, we use the same classifier as in Section~\ref{subsec: baseline_exps} which has only seen images without any shadow. The testing data consists of shadowy images with 3 different attributes: different pole widths to simulate different shadow sizes, different pole transparency to simulate different shadow intensity, and different rotations and translations to simulate random shadow placement. To evaluate each attribute, we take all combinations of other attributes and average the results. For example, to evaluate the impact of one pole width, we combine all pole transparency and placement at that pole width and average the results. We use the same evaluation metrics as in the previous section: top1 accuracy and the breakdown points. 

\subsubsection{Results and discussion}
\begin{figure}[htb!]
\centering

\begin{subfigure}[t]{0.85\linewidth}
    \includegraphics[width=\textwidth]{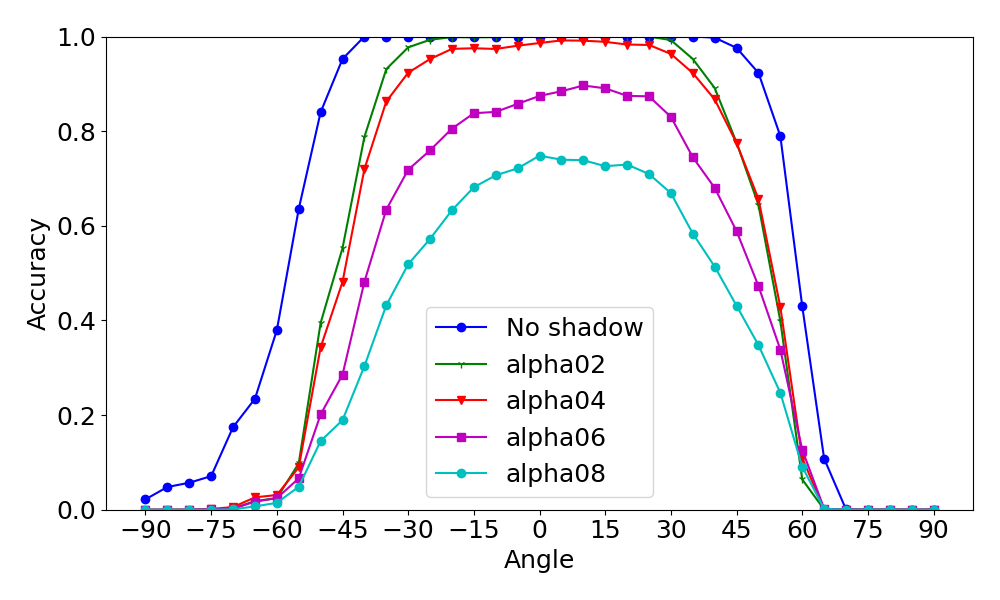}
    \caption{Top1 acc. of rub back with varied pole transparency.}
\end{subfigure}
\vspace{.1cm}
\begin{subfigure}[t]{0.85\linewidth}
    \includegraphics[width=\textwidth]{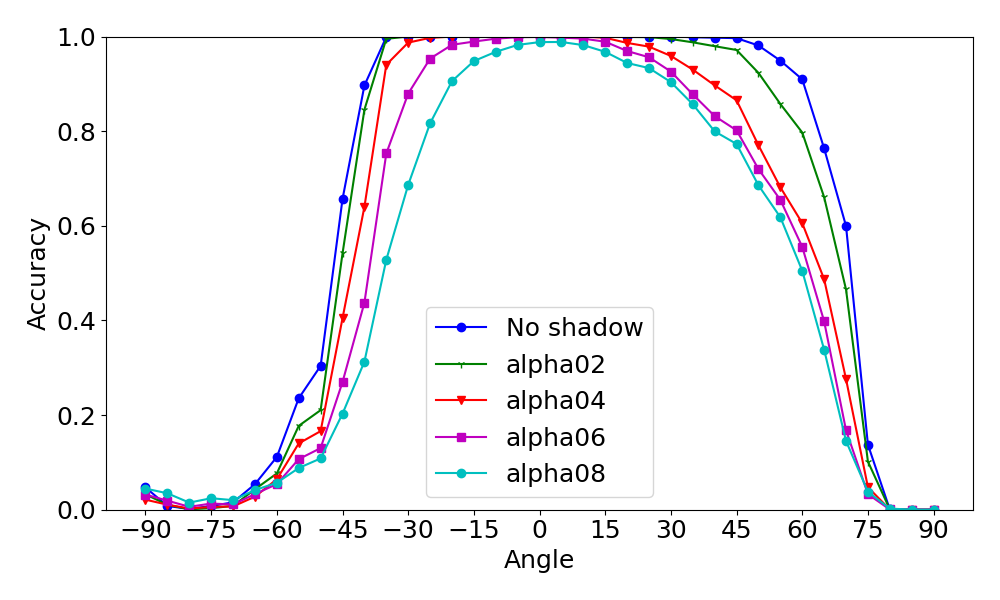}
    \caption{Top1 acc. of rub thumb with varied pole transparency.}
\end{subfigure}%

\caption{Top1 accuracy of varied pole transparency.}
\label{fig:shadow_alpha_results}
\end{figure}

\begin{figure}[htb!]
\centering

\begin{subfigure}[t]{0.47\linewidth}
    \includegraphics[width=\textwidth]{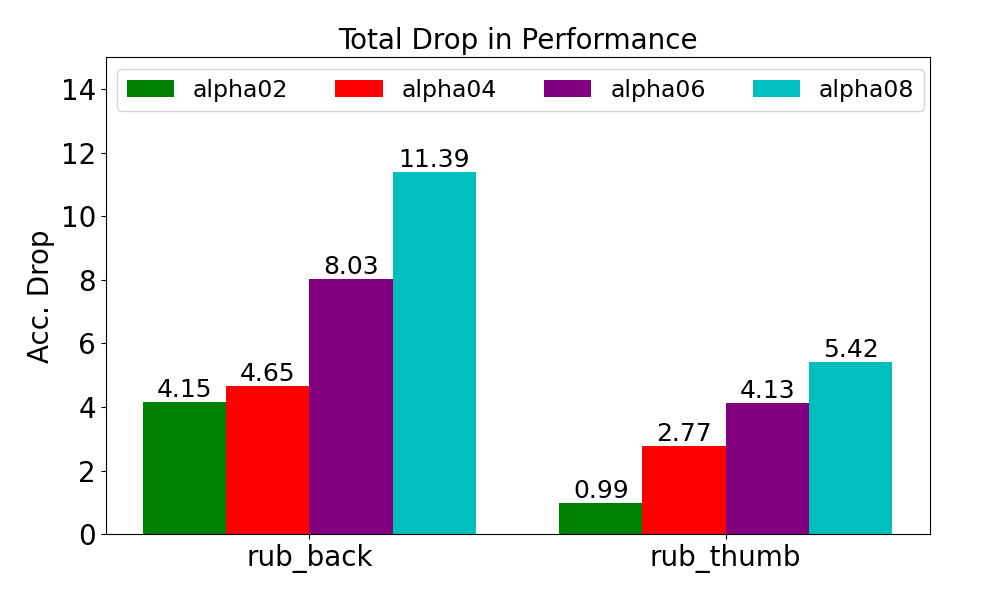}
    \caption{Total top1-acc. drop of varied pole transparency.}
    \label{subfig: alpha_drop_acc}
\end{subfigure}%
\hspace{.1cm}
\begin{subfigure}[t]{0.47\linewidth}
    \includegraphics[width=\textwidth]{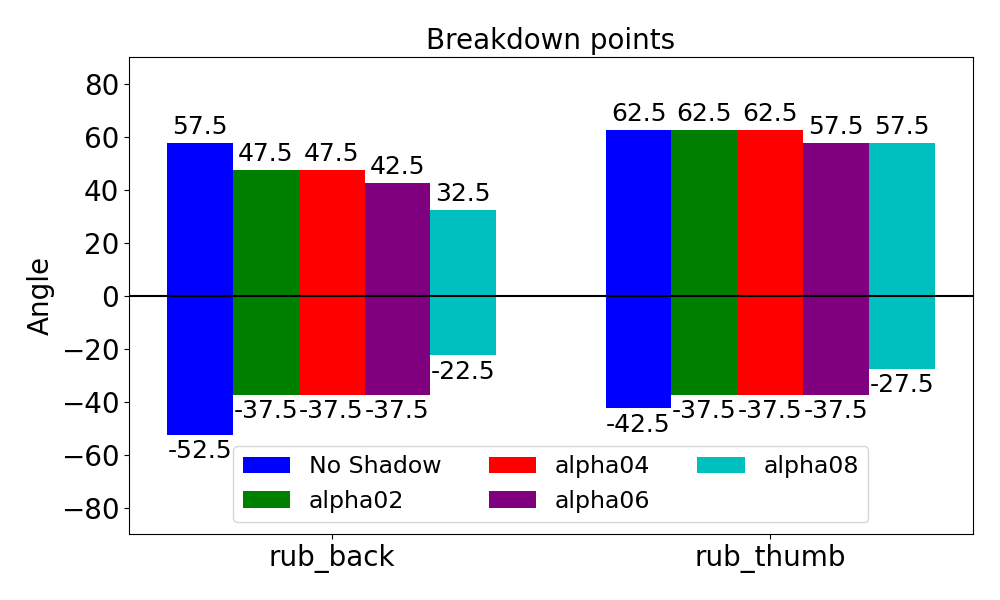}
    \caption{Breakdown points of varied pole transparency.}
    \label{subfig: shadow_alpha_bdpts}
\end{subfigure}%

\caption{Pole transparency's impact on performance.}
\label{fig:shadow_alpha_bdpts}
\end{figure}

Figure~\ref{fig:shadow_alpha_results} shows the overall top1 accuracy for varying pole transparency for the two actions. 
In addition, Figure~\ref{subfig: alpha_drop_acc} shows that the overall drop of top1 accuracy increases as the shadow becomes more intense. Also, the overall performance drop is much larger for rub back than for rub thumb. 

From the results in Figure~\ref{subfig: shadow_alpha_bdpts} that show the breakdown points, we can see that as shadow intensity increases, performance breaks down earlier. This shift in breakdown points is again more noticeable for rub back than rub thumb. This could be due to the fact that rub thumb contains a more unique pose, making it less susceptible to distribution shift like shadows.

\begin{figure}[htb!]
\centering

\begin{subfigure}[t]{0.47\linewidth}
    \includegraphics[width=\textwidth]{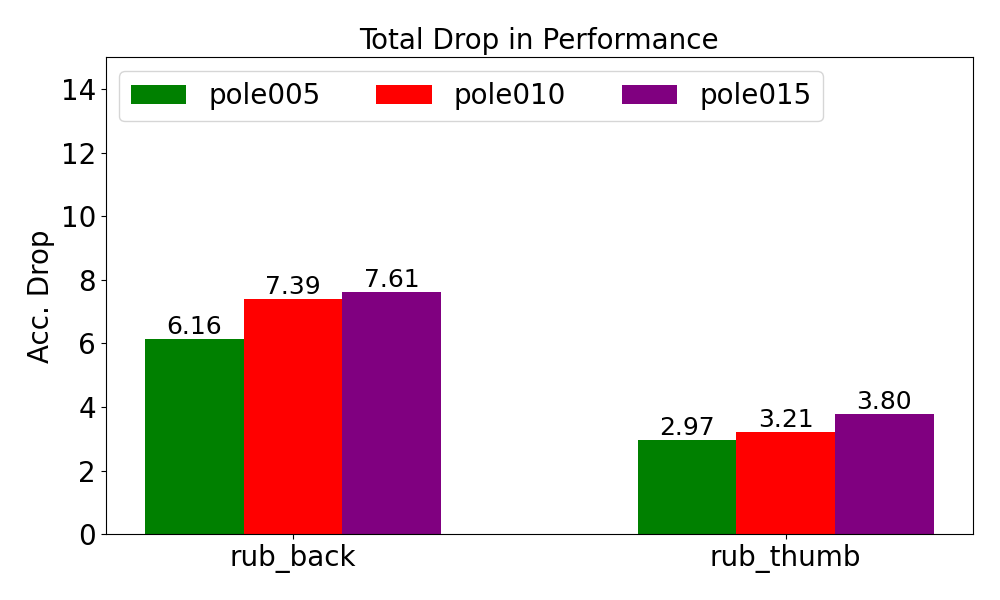}
    \caption{Total top1-acc. drop of varied pole widths.}
    \label{subfig: pole_drop_acc}
\end{subfigure}%
\hspace{.1cm}
\begin{subfigure}[t]{0.47\linewidth}
    \includegraphics[width=\textwidth]{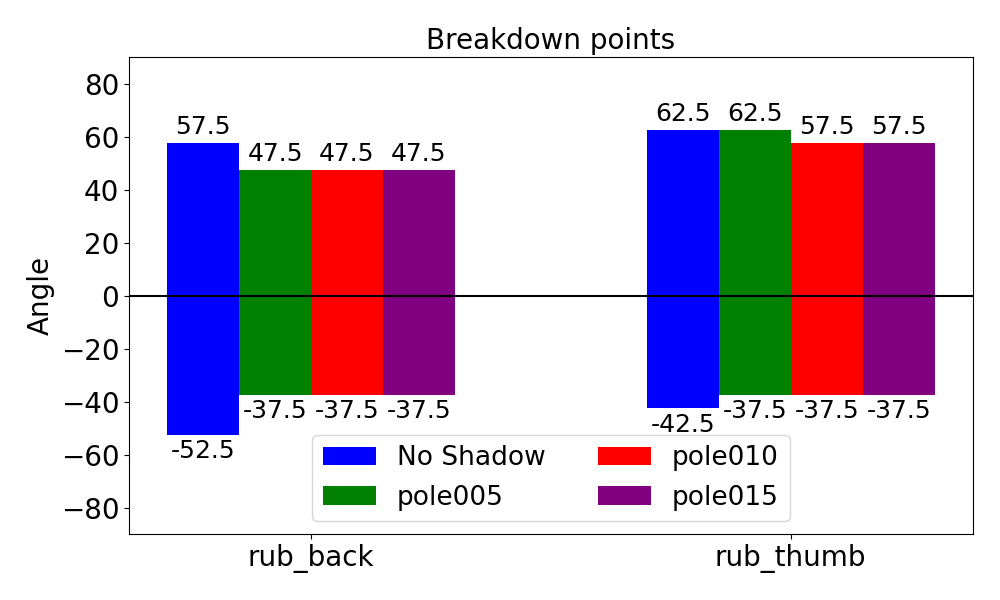}
    \caption{Breakdown points of varied pole widths.}
    \label{subfig: pole_bdpts}
\end{subfigure}%

\caption{Pole width's impact on performance.}
\label{fig:shadow_width_bdpts}
\end{figure}

For results of varied pole widths, we omit the top1 accuracy plots and directly show the overall accuracy drop and the breakdown points. 
Figure~\ref{subfig: pole_drop_acc} shows the total drop of top1 accuracy for varied pole widths. Similar to the pole transparency scenario, we see an increase in total accuracy drop as pole width increases. However, the change in total accuracy drop is much smaller than in the case of pole transparency, which indicates that shadow size has less impact on overall performance than shadow intensity. 
Figure~\ref{subfig: pole_bdpts} shows the breakdown points with varied pole width. As can be seen from the results, the breakdown points shift earlier but only slightly. This also shows that shadow intensity plays a bigger role in impacting performance than shadow size. 



\vspace{-.15in}

\begin{figure}[htb!]
    \centering
    \includegraphics[width=0.7\linewidth]{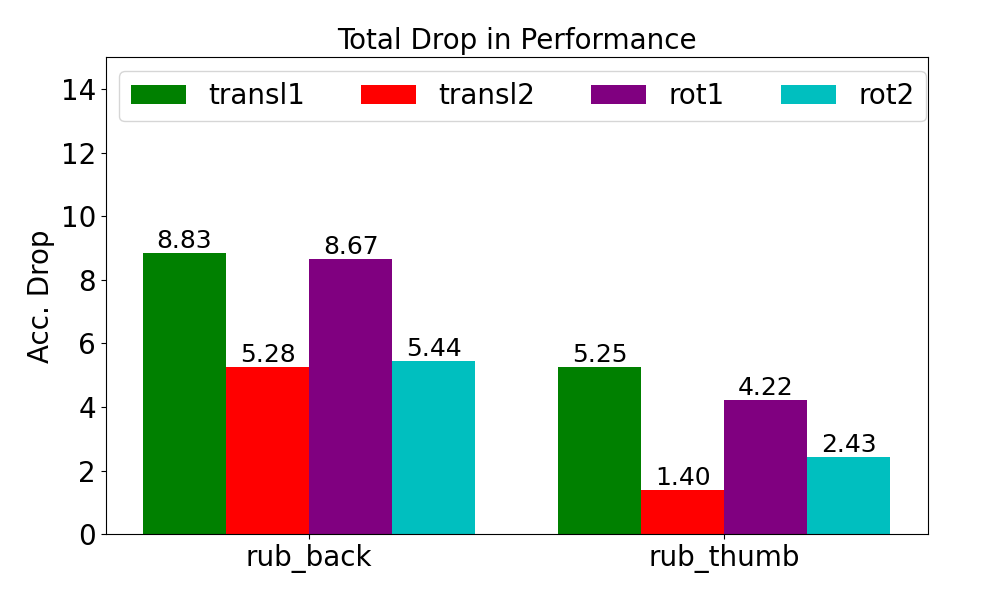}
    \caption{Total top1-acc. drop of varied pole placements.}
    \label{fig:shadow_loc_rot_top1}
\end{figure}




Figure~\ref{fig:shadow_loc_rot_top1} shows the overall drop in top1 accuracy for varied pole placements. As can be seen from Figure~\ref{fig:shadow_rot_loc}, a shadow placed at translation 1 covers more hand regions than at translation 2. Also, a shadow at rotation 1 covers more hand regions than at rotation 2. We observe more performance drop when the shadow covers more hand regions. We omit the figure for breakdown points to save space. The breakdown points follow the same trend as the top1 accuracy, where higher drop translates to earlier breakdown points.
Whether shadow is placed over important regions of objects has significant impact on the breakdown points and overall performance. 


\subsection{Which additional training poses mitigate the pose breakdown points?}
\subsubsection{Experimental setup}
We explore strategies to mitigate the significant drop in performance caused by pose-induced distribution shift by simply adding specific additional poses into the training set. Keeping the 0-degree training data fixed as in the previous two sets of experiments, we discretely sample and add hand poses from angles other than 0 to training data. The goal is to investigate which angle provides the most gain to performance and also best mitigates the pose-induced breakdown points. 
Hand angles are sampled symmetrically for positive and negative angles. We sample based on the range from 5 to 90 degrees with a 5-degree step, and we train separate models for each additional angle. The training set now consists of an equal number of images from 0 degree, $+x$ degree, and $-x$ degree. Model architecture and other hyperparameters are kept the same to ensure a fair comparison. 

\subsubsection{Results and discussion}

\begin{figure}[htb!]
\centering
\begin{subfigure}[t]{0.47\linewidth}
    \includegraphics[width=\textwidth]{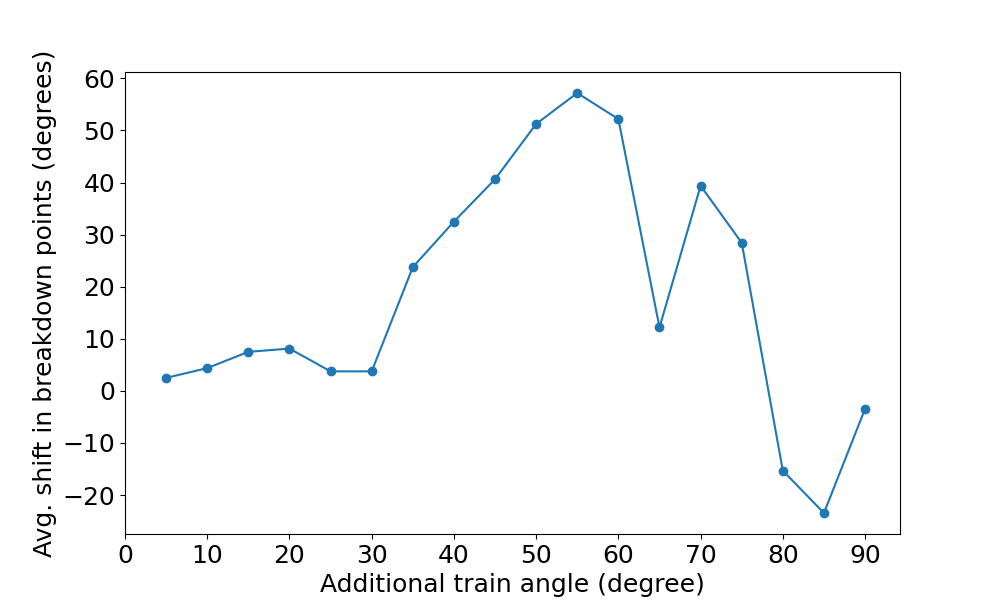}
    \caption{Average change in breakdown points for each additional training angle.}
\end{subfigure}%
\hspace{.1cm}
\begin{subfigure}[t]{0.47\linewidth}
    \includegraphics[width=\textwidth]{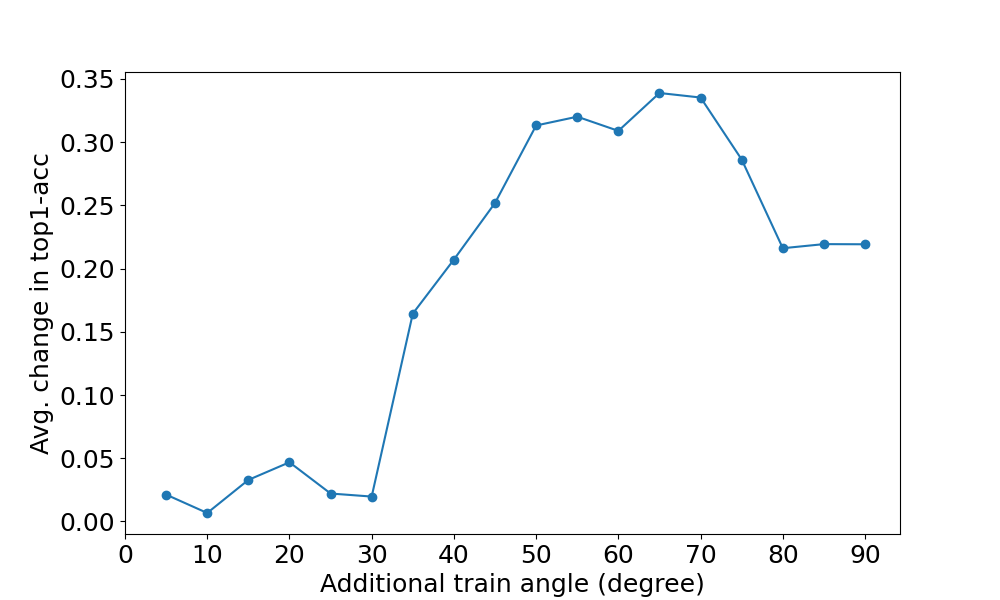}
    \caption{Average change in top1 acc. for each additional training angle.}
\end{subfigure}%

\caption{Results with additional training data.}
\label{fig:addtl_angle}
\end{figure}

Figure~\ref{fig:addtl_angle} shows the average changes in breakdown points and in top1 accuracy for each additional training angle. For both of these changes, the larger the better. As can be seen from the figures, additional data with small deviations from 0-degree yields no significant improvement, as shown by the parts of the plots from 5 to 30 degree. Moreover, additional data with large deviations from 0-degree results in higher top1 accuracy but worse breakdown points, as indicated by the 80 to 90 degree regions of the plots. This is due to the top1 accuracy falling below the breakdown point threshold at an angle closer to 0, which is caused by training the model with significantly different poses that belong to the same action. Overall, the optimal choices of additional training pose fall into the range between 50 and 60 degrees, where breakdown points are significantly mitigated and overall performance is largely improved.

\section{Conclusion}
\label{sec:conclusion}
In conclusion, in this paper we investigate the impact of pose and shadow on a classifier's performance. We generate synthetic data with varied hand poses and shadow conditions to introduce distribution shift in a controlled manner. Then, we quantitatively evaluate how pose and shadow impacts a classifier by studying the breakdown points. Results show that pose causes model performance to drop sharply after the breakdown points. Also, larger and heavier shadow will cause the system to perform worse overall and start degrading in its performance earlier. In particular, results indicate that shadow intensity has more impact on performance than shadow size. 
Furthermore, we explore a simple yet effective approach to mitigate the pose-induced breakdown points by using additional training angle. Results show that hand poses around 50-60 degrees of rotation are optimal choices for alleviating the breakdown points. Future research includes quantitatively measuring the pose- and shadow-induced distribution shift using available measures and bridging the gap between real and synthetic datasets.


\bibliographystyle{IEEEtran}
\bibliography{ref}

\end{document}